\documentclass[10pt,twocolumn,letterpaper]{article}

\usepackage{cvpr}              % To produce the CAMERA-READY version

\usepackage{multirow}
\usepackage{color} % 基本颜色支持
\usepackage{xcolor} % 扩展颜色支持
\usepackage{colortbl}
\usepackage{placeins}
\usepackage[accsupp]{axessibility}  % Improves PDF readability for those with disabilities

\definecolor{cvprblue}{rgb}{0.21,0.49,0.74}
\usepackage[pagebackref,breaklinks,colorlinks,allcolors=cvprblue]{hyperref}

\def\paperID{   } % *** Enter the Paper ID here
\def\confName{CVPR}
\def\confYear{2026}
\def\secref#1{Sec.~\ref{#1}}
\def\figref#1{Fig.~\ref{#1}}
\def\tabref#1{Tab.~\ref{#1}}
\def\eqref#1{Eq.~(\ref{#1})}

\title{TAPFormer: Robust Arbitrary Point Tracking via Transient Asynchronous Fusion of Frames and Events}

\author{Jiaxiong Liu$^{1}$\and Zhen Tan$^{1}$\and Jinpu Zhang$^{1}$\and Yi Zhou$^{2}$\and Hui Shen$^{1}$\and Xieyuanli Chen$^{1}$\hspace{1.0cm} Dewen Hu$^{1}$\\
{ $^1$National University of Defense Technology \hspace{1.0cm} $^2$Hunan University}
}
\begin{document}
\maketitle
\begin{abstract} 
Tracking any point (TAP) is a fundamental yet challenging task in computer vision, requiring high precision and long-term motion reasoning. 
Recent attempts to combine RGB frames and event streams have shown promise, yet they typically rely on synchronous or non-adaptive fusion, leading to temporal misalignment and severe degradation when one modality fails. 
We introduce TAPFormer, a transformer-based framework that performs asynchronous temporal-consistent fusion of frames and events for robust and high-frequency arbitrary point tracking.
Our key innovation is a Transient Asynchronous Fusion (TAF) mechanism, which explicitly models the temporal evolution between discrete frames through continuous event updates, bridging the gap between low-rate frames and high-rate events. In addition, a Cross-modal Locally Weighted Fusion (CLWF) module adaptively adjusts spatial attention according to modality reliability, yielding stable and discriminative features even under blur or low light.
To evaluate our approach under realistic conditions, we construct a novel real-world frame-event TAP dataset under diverse illumination and motion conditions.
Our method outperforms existing point trackers, achieving a $28.2\%$ improvement in average pixel error within threshold. 
Moreover, on standard point tracking benchmarks, our tracker consistently achieves the best performance. Project website: \url{tapformer.github.io}

\end{abstract}    
\section{Introduction}
\label{sec:intro}

\begin{figure}[t]
   \centering
   \setlength{\abovecaptionskip}{-0.05cm}
   % \vspace{-0.7cm}
   \includegraphics[width=0.46\textwidth]{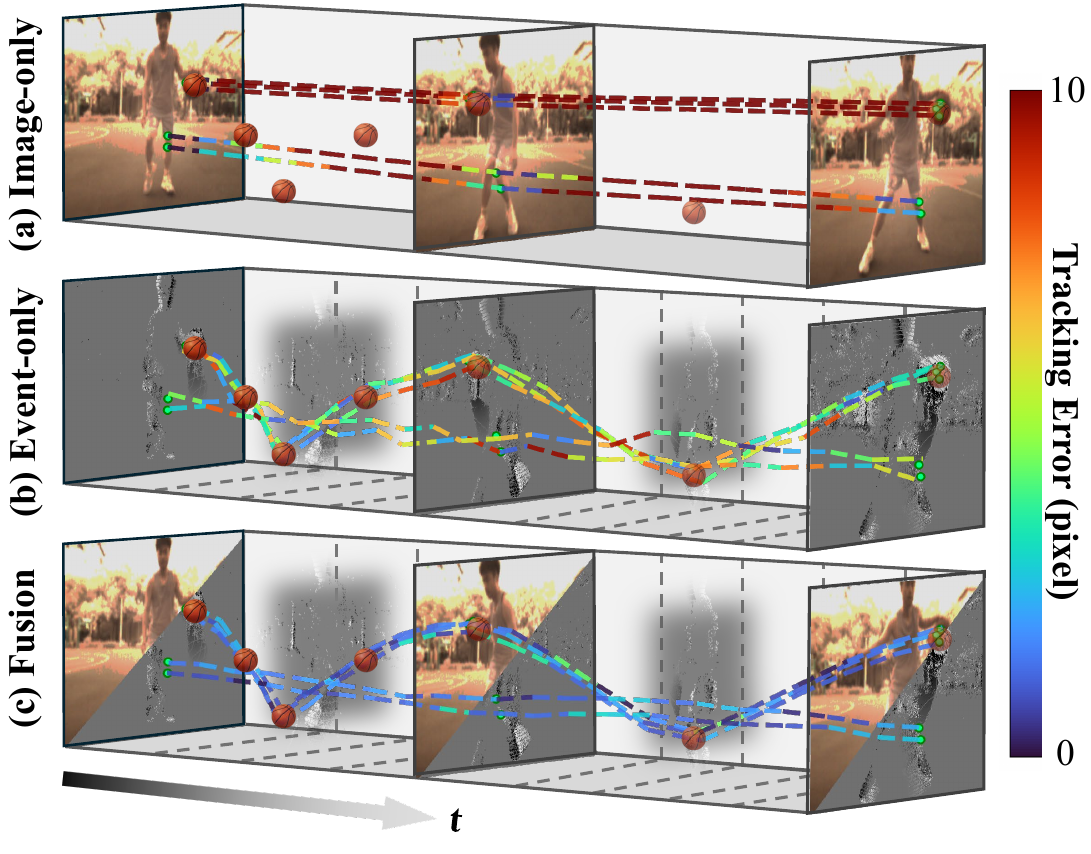} 
   \caption{A qualitative comparison of tracking performance in our dataset shows that (a) frame-based tracking suffers from insufficient temporal information, and (b) event-based tracking fails to capture fine spatial details. In contrast, our (c) fusion approach can recover long-term, high-accuracy point trajectories. The rightmost color bar visualizes the per-point tracking error (in pixels).}
   \label{fig:example}
   \vspace{-0.7cm}
\end{figure}

Tracking any point (TAP), which estimates the motion trajectory of any point in a video, is a fundamental task in computer vision. A robust tracker capable of maintaining accurate trajectories under diverse conditions, from bright daylight to low-light nighttime scenes and from slow to fast motions, is important for applications such as augmented reality and autonomous systems~\cite{chen2024leap, vecerik2024robotap}. However, existing methods relying on conventional frame-based cameras often experience severe degradation under challenging conditions, due to motion blur, overexposure, or limited frame rates that fail to capture rapid dynamics~\cite{harley2025alltracker, cho2024local, wang2023tracking, karaev2025cotracker3}.

Event cameras, in contrast, asynchronously record per-pixel brightness changes with microsecond temporal precision, offering wide dynamic range. These properties make them suitable for high-speed or high-contrast environments.
However, event data are inherently coupled with motion, meaning the generated event stream depends on both the scene and the camera motion.
Even the same scene can produce drastically different event patterns under varying motions, making it difficult to form consistent feature maps across motion conditions. 
Moreover, in slow-motion or static scenes, the event stream becomes sparse, and the lack of texture and color further limits tracking performance.

These complementary strengths and weaknesses naturally motivate fusion-based approaches~\cite{conf_cvpr_MessikommerFG023, wang2021joint}. In biological vision, robust perception is achieved through the cooperation of parallel visual pathways. The ventral visual stream mainly processes static attributes such as color and texture, while the dorsal stream encodes motion and spatial relationships~\cite{yang2024vision}.  
Analogously, RGB and event cameras provide complementary visual information. The former excels at capturing detailed spatial structure, while the latter encodes fine-grained temporal dynamics. Inspired by this analogy, we develop a unified fusion framework that integrates both modalities to achieve stable, high-precision arbitrary point tracking across diverse real-world conditions.

To achieve this goal, two major challenges should be addressed.
The first challenge lies in the frequency mismatch between image frames and event data. 
Directly fusing asynchronous event streams with the most recent image frame can result in severe spatial misalignment due to the time lag between the two modalities. 
Existing fusion methods~\cite{tomy2022fusing, liu2021attention} often circumvent this problem by downsampling event data to match frame rates, sacrificing temporal fidelity and limiting potential gains.
Although asynchronous fusion techniques have been explored in related tasks~\cite{gehrig2024low, zhang2023frame}, they are not well suited for TAP, as it requires pixel-level precision, long-term consistency, and robust handling of fine-grained features.
To address this, we introduce a Transient Asynchronous Fusion (TAF) mechanism. Instead of aligning event streams to discrete frames, TAF treats the scene as a temporally continuous latent representation. Each time a frame arrives, the fusion state is initialized by combining frame and recent events; between frames, this representation is continuously updated by incoming events through a lightweight cross-attention updater. This formulation explicitly models the temporal evolution of visual information, enabling high-frequency feature updates far beyond the frame rate.

The second challenge lies in constructing robust cross-modal feature representations under challenging conditions.
In realistic scenes, the reliability of image and event modalities fluctuates spatially and temporally. Cross-modal Locally Weighted Fusion (CLWF) module introduces an adaptive local attention mechanism where event tokens query spatial neighborhoods of frame tokens, learning to emphasize the modality that is currently more informative. This results in robust, spatially coherent features even when one modality degrades due to motion blur or sparsity.
Moreover, once frame and event inputs are tokenized and fused, both scale and temporal information can be naturally incorporated into our architecture.
In this way, these modules establish a temporally consistent, modality-adaptive representation that bridges the temporal gap between asynchronous sensors while maintaining fine spatial precision. 

Beyond algorithmic contributions, we also introduce two datasets to systematically study multimodal TAP. We built a new multimodal tracking benchmark that includes a synthetic frame-event training dataset and manually annotated real-world test sequences.
Comprehensive experiments are conducted on both TAP and feature point tracking tasks.
Across all benchmarks, our method consistently achieves the best performance among frame-based, event-based, and fusion-based trackers, demonstrating its strong generalization and robustness under diverse real-world conditions. 
The contribution of our work consist of:
\begin{itemize}
\item{We propose a unified framework that fully exploits the complementary strengths of RGB frames and event streams, enabling robust arbitrary point tracking under diverse challenging real-world conditions.}
\item{The first real-world TAP benchmark covering challenging conditions with synchronized frame–event data.}
\item {State-of-the-art performance across multiple datasets in TAP and feature point tracking tasks, demonstrating the effectiveness and generality of our approach.}
\end{itemize}

By shifting frame–event fusion from synchronous aggregation to asynchronous temporal modeling, TAPFormer offers a new perspective for building continuous, event-driven perception systems.

\section{Related Work}
\label{sec:Related Work}

\subsection{Frame-Based Point Tracking Methods}
TAP task has been revisited and reformulated as a particle video problem by querying trajectories across the entire image sequence~\cite{harley2022particle}, which handles occlusions by leveraging temporal context. TAP-Vid~\cite{doersch2022tap} further provides both synthetic training data and standardized evaluation benchmarks. PointOdyssey~\cite{zheng2023pointodyssey} also provides a large-scale synthetic dataset for long-term TAP.
CoTracker~\cite{karaev2024cotracker} models the spatial correlations among target points by leveraging their physical coherence and jointly tracks them within a sliding temporal window. Inspired by DETR~\cite{carion2020end}, TAPTR~\cite{li2024taptr} employs a transformer-based architecture as an iterative optimizer and achieves strong performance on benchmark datasets. CoTracker3~\cite{karaev2025cotracker3} further simplifies the architecture by removing redundant components, achieving higher efficiency with minimal accuracy loss. Chrono~\cite{kim2025exploring} introduces temporal adapters between the transformer blocks of DINOv2~\cite{oquab2023dinov2}, demonstrating that temporal coherence plays a key role in feature extraction.

Another related line of research is optical flow estimation~\cite{teed2020raft, ilg2017flownet}, which focuses on dense correspondence estimation between consecutive frames. However, such methods often suffer from drift in long-term tracking. Despite recent advances in frame-based tracking, the inherent limitations of standard cameras, such as fixed frame rates and limited dynamic range, restrict their applicability under fast-motion or low-light conditions.

\subsection{Event-Based Point Tracking Methods}
Early event-based tracking approaches relied on geometric or optimization-based models\cite{conf_icra_ZhuAD17, conf_bmvc_AlzugarayC20, journals_ral_AlzugarayC18}. For instance, Zhu et al.~\cite{conf_icra_ZhuAD17} treat event streams as point clouds and estimate trajectories using ICP algorithms. HASTE~\cite{conf_bmvc_AlzugarayC20} tracks features at the event level by hypothesizing multiple motion patterns and selecting the best match through template comparison. 
To enhance the performance and robustness of trackers, data-driven approaches have been proposed~\cite{burkhardt2025superevent, huang2023eventpoint, han2025mate, messikommer2025data}. However, these methods are limited to predefined feature points and cannot generalize to arbitrary point tracking.

More recently, ETAP~\cite{hamann2025etap} adapted trajectory optimization networks from the TAP task to the event domain and proposed an event-based TAP synthetic training dataset. 
MATE~\cite{han2025mate} introduces a motion-augmented approach that significantly enhances the temporal consistency and robustness of event-based point tracking.

Despite these advances, event-based tracking often fails in low-motion or static scenes due to sparse event generation. Moreover, the lack of color and texture detail limits the tracking precision of event-only methods, making them less reliable than frame-based approaches in most scenarios.

\subsection{Multimodal Fusion for Events and Frames}
To achieve more robust point tracking, several studies have explored the integration of frame and event data to leverage their complementary strengths: rich textures and stable appearance from images, and high temporal resolution and dynamic range from events~\cite{journals_ijcv_GehrigRGS20, conf_cvpr_MessikommerFG023, liu2025tracking, conf_icra_WangCYYXY24, journals_ral_WangYYYX24, wan2025event}. For example, EKLT~\cite{journals_ijcv_GehrigRGS20} uses grayscale images as templates and matches them with brightness increment images generated from event streams to perform feature point tracking. Other works adopt data-driven approaches, significantly improving performance under challenging conditions. 
While the recent FETAP~\cite{liu2025tracking} introduces the first TAP model that fuses images and events, achieving substantial performance gains, its training on optical flow data prevents it from determining point occlusion.

Fusion frames and events have also been investigated in other vision tasks such as object tracking~\cite{gehrig2024low, zhang2023frame, liu2024enhancing, li2023sodformer}, line segment detection~\cite{yu2023detecting}, and depth estimation~\cite{gehrig2021combining}. 
However, most existing methods are either constrained by frame rates or suffer from degradation when one modality is unreliable~\cite{tomy2022fusing}. 
Moreover, these approaches are not directly applicable to TAP, which demands pixel-level precision, long-term consistency, and robust per-point feature representations that remain stable under irregular motion.

To address these limitations, we propose a unified asynchronous fusion framework featuring two key modules. 
First, a transient feature update mechanism integrates RGB and event information to construct a temporally consistent representation when an image arrives and continuously updates it with incoming events between frames, enabling high-frequency tracking beyond standard frame rates. 
Second, a cross-modal locally weighted fusion module adaptively selects reliable regions from each modality and fuses them at a local scale, enhancing robustness under motion blur, occlusion, and illumination changes. 
Together, these designs yield a stable and temporally coherent representation that enables precise and robust arbitrary point tracking under diverse real-world conditions.

\section{Method}
\label{sec: Method}
\begin{figure*}[t!]
    % \vspace{0.15cm}
    \centering
    \includegraphics[width=1\linewidth]{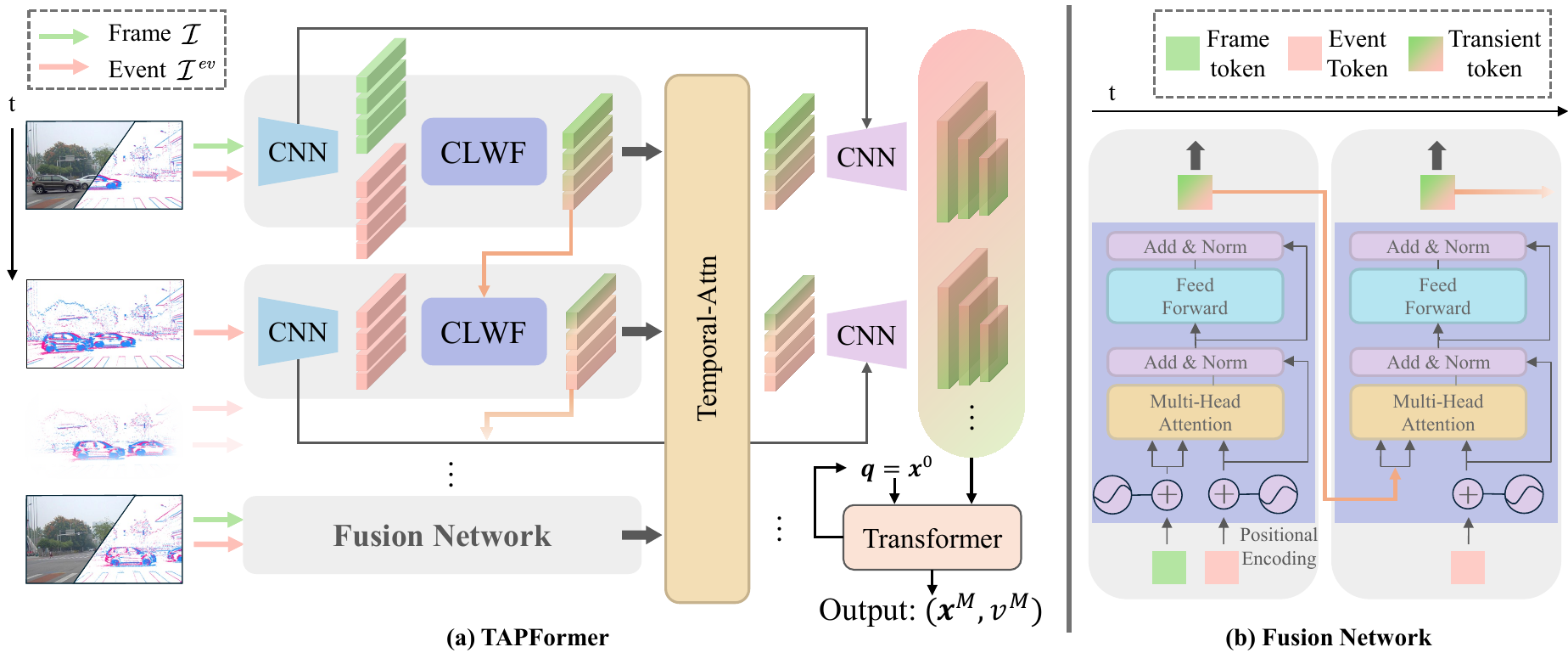}
    \vspace{-0.7cm}
    \caption{TAPFormer overview. (a) The overall framework: frames and events are fused by the transient asynchronous fusion mechanism and Cross-Modal Local Weighted Fusion modules to produce high-frequency transient features, refined by temporal attention and decoded into multi-scale fusion features. The resulting features, together with the initial query points position \(\textbf{q}\), are fed into a transformer-based optimization module to iteratively predict tracking trajectories \(\textbf{x}\) and occlusion states \(v\). \(M\) denotes the number of iterations. (b) The fusion network: image and event tokens are integrated by local weighted cross-attention to construct and update transient representations.}
    \vspace{-0.6cm}
    \label{fig:overview}
\end{figure*}
\subsection{Problem Definition}
Our TAP framework fuses RGB and event data. Event cameras asynchronously record pixel-wise brightness changes, triggering an event whenever the logarithmic intensity variation exceeds a threshold \(c\in \mathbb{R}^+ \).  
Each event \( e_k \doteq (\boldsymbol{x}_k, t_k, p_k) \) includes pixel location \( \boldsymbol{x}_k=(x_k, y_k) \), a microsecond timestamp \( t_k \), and polarity \( p_k \in \{-1, +1\} \) for the sign of brightness change.
The event-triggering condition can thus be formally expressed as:
\begin{equation}
    \Delta L(\boldsymbol{x}_k, t_k) = L(\boldsymbol{x}_k, t_k) - L(\boldsymbol{x}_k, t_k - \Delta t_k) = p_k c,
\end{equation}
where \( L = \log(I) \) denotes the logarithmic intensity of image \( I \in \mathbb{R}^{H \times W \times 3}\). \(\Delta t_k\) denotes the short interval between two consecutive events at the same pixel. Consequently, the pixel intensity at time \( t_k \) can be recovered from its previous intensity using the event polarity and threshold:
\begin{equation}
    I(\boldsymbol{x}_k, t_k) = \exp(p_k c) \cdot I(\boldsymbol{x}_k, t_k - \Delta t_k).
\label{eq:generation model}
\end{equation}

Next, we define the TAP task that fuses image and event data. For simplicity, we represent the entire target process using a single query point here. Given a time interval \(\tau = [\tau_s, \tau_e]\in \mathbb{R}\ \),\(\ \tau_s \le \tau_e\), we observe a sequence of events \(\mathcal{E} = \{e_k \mid  t_k \in  \tau\}\) and a sequence of images \(\mathcal{I} = \{I_t\}_{t=0}^{N^i}\), where \(N^i \in \mathbb{N^+}\) is the number of image frames.
We aim to track a query point \(\mathbf{q} = (t_q, x, y) \in \mathbb{R}^3\), 
initialized at time \(t_q \in {\tau}\), and estimate its trajectory 
\(\boldsymbol{x}_{\tau_t} = (x_{\tau_t}, y_{\tau_t}) \in \mathbb{R}^2\) and visibility 
\({v}_{\tau_t} \in \{0,1\}\) over discrete time steps \(\{\tau_t\}_{t=0}^{N^e-1} \subset \tau \), 
where \(N^e \in \mathbb{N^+}\) denotes the number of event frames, and 
1 indicates visibility while 0 denotes occlusion.

For efficiency and noise robustness, the tracker operates at all discrete query times \( \{\tau_t\}_{t=0}^{N^e-1} \),  whose update frequency \( f_e = \frac{N^e}{\tau_e-\tau_s} \) 
(typically 100–200\,Hz) is much higher than the image frame rate \( f_i = \frac{N^i}{\tau_e-\tau_s} \) (20–30\,Hz). 
In principle, using our transient asynchronous fusion mechanism, the tracker can estimate the query point location at arbitrary query times $\tau_t$. At each discrete step \(\tau_t\), the set of events in the corresponding time bin is collected:
\begin{equation}
E_t = \{e_k \mid \tau_{t-1} < t_k \le \tau_t\} \subset \mathcal{E},
\end{equation}
and convert them to event representations \( I^{ev}_t = \mathcal{F}(E_t) \in \mathbb{R}^{H \times W \times B} \). \(B\) is the channel number. \(\mathcal{F}\) is a maximal timestamp version of Stacking Based on Time (SBT)~\cite{wang2019event}.

In summary, our goal is to use high-frequency event representations \(\mathcal{I}^{ev}\), the images \(\mathcal{I}\), and the initial query state \(\mathbf{q}\), to estimate the pixel trajectory \(\boldsymbol{x}_{\tau_t}\) and visibility state \(v_{\tau_t}\) of the query point at each \(\tau_t\), as shown in ~\figref{fig:overview}.

\subsection{Transient Asynchronous Fusion}

A core challenge in frame-event fusion arises from their distinct temporal sampling characteristics: image frames are sparsely captured at low rates, whereas events provide fine-grained, asynchronous temporal updates. To bridge this temporal-frequency gap, we propose a Transient Asynchronous Fusion (TAF) mechanism that establishes and continuously updates a unified transient representation.

The Event-based Double Integral (EDI) model~\cite{pan2019bringing, lin2023fast} describes a blurred image \(\tilde{\mathbf{B}}\) as the temporal integration of a sequence of latent sharp images \(\tilde{\mathbf{L}}(t')\). Formally, it can be expressed as
\begin{equation}
\tilde{\mathbf{L}}(t') = \tilde{\mathbf{B}} - 
\log\!\left(
\frac{1}{T}\!\int_{t'-T/2}^{t'+T/2}\!
\exp(c\,\mathbf{E}(t))\,dt
\right),
\label{eq:edi}
\end{equation}
which suggests that the observed image represents an integration of continuous latent intensity changes over time, modulated by the event data $\mathbf{E}(t)$.
Building on this observation, we consider that a conventional frame and the corresponding events within its exposure interval jointly encode complementary temporal information, enabling the recovery of an instantaneous scene state.
Following this idea, each incoming frame $I_t$ is fused with events generated during its exposure period, defined as a short window $\mathcal{W}_t=(t-\delta,t]$. The exposure duration can be treated as a constant hyperparameter when the actual exposure time is unavailable. Fusion is performed through a cross-modal local weighted fusion (CLWF) module (detailed in ~\secref{sec_:clwf}):
\begin{equation}
\mathcal{R}_t = 
\mathcal{G}\big(\Phi_I(I_t),\,\Phi_E(\mathcal{F}(\mathcal{E}\cap\mathcal{W}_t))\big),
\label{eq:taf_init}
\end{equation}
where $\Phi_I$ and $\Phi_E$ are modality-specific encoders. $\mathcal{G}$ denotes the CLWF operator that balances modality contributions in local spatial regions. $\mathcal{R}_t$ is transient representation.

After the initial fusion, $\mathcal{R}_t$ is asynchronously refined as new batches of events arrive. 
From \eqref{eq:generation model}, based on the event generation model~\cite{gallego2020event}, the image intensity at time $t$ can, in theory, be reconstructed from the intensity in a previous frame and the accumulated events in between.
However, this model is highly sensitive to sensor noise and hardware non-idealities. 
Instead of directly reconstructing pixel intensities, we apply an event-driven residual refinement mechanism through a cross-attention-based updater:
\begin{equation}
\mathcal{R}_{t+\Delta}
\leftarrow
\mathcal{U}\big(\mathcal{R}_{t+\Delta-1},\,\Phi_E( \mathcal{F}(E_{t+\Delta}))\big),
\label{eq:taf_update}
\end{equation}
where $\mathcal{U}$ integrates fine-grained event cues while preserving spatial consistency from the recent frame. 
This design maintains temporal fidelity at event-rate updates, enabling smooth and accurate tracking in high-speed conditions.

\subsection{Cross-modal Local Weighted Fusion Module}
\label{sec_:clwf}

% \noindent\textbf{Cross-modal Local Weighted Fusion Module.}
The Cross-Modal Local Weighted Fusion (CLWF) module acts as the key component in constructing the transient representation. It adaptively integrates image and event tokens, converted by a specific encoder $\Phi$, within local regions in a modality-aware manner. The goal is to enhance local reliability by assigning higher attention weights to the more informative modality in each spatial neighborhood.

Given the image tokens $\Phi_I(I_t)\in\mathbb{R}^{N\times d}$ and event tokens $\Phi_E(\mathcal{F}(E\cap\mathcal{W}_t))\in\mathbb{R}^{M\times d}$, CLWF performs a localized cross-attention operation. Each event token serves as a query that gathers information from image tokens within its spatial neighborhood $\mathcal{N}(j)$. The fusion weights are:
\begin{equation}
A_{j,i} = 
\frac{\exp\big(\langle q_j, k_i\rangle/\sqrt{d} + \mathcal{M}_{j,i}\big)}
{\sum_{i'\in\mathcal{N}(j)} 
\exp\big(\langle q_j, k_{i'}\rangle/\sqrt{d} + \mathcal{M}_{j,i'}\big)},
\label{eq:local_attn_final}
\end{equation}
where $\mathcal{M}_{j, i}$ is an optional learned locality bias that encourages spatially adjacent tokens to have a stronger influence.  
Each event token is then updated through a residual cross-attention fusion: \(\mathbf{f}^{tra}_j = \mathbf{f}^E_j + 
\sum_{i\in\mathcal{N}(j)} A_{j,i}\,v_i,\)
% \begin{equation}
% \mathbf{f}^{tra}_j = \mathbf{f}^E_j + 
% \sum_{i\in\mathcal{N}(j)} A_{j,i}\,v_i,
% \label{eq:fused_token_final}
% \end{equation}
which preserves the event-driven temporal dynamics while injecting complementary image cues.

To enhance spatio-temporal coherence, the fused transient tokens $(\mathcal{R}_t = \{\mathbf{f}^{tra}_j\}$ are processed by a temporal self-attention block that aggregates information across adjacent time steps. Positional (spatial and temporal) embeddings are added before attention to preserve ordering information. During decoding, the resulting spatio-temporal token representation is progressively upsampled with skip connections to the corresponding encoder features, thereby producing multi-scale fused feature maps that integrate fine-grained temporal detail with global semantic context.

Finally, these multi-scale fused feature maps, together with the initial query point information, are fed into a transformer-based optimizer to iteratively estimate the query point trajectory and visibility.

\section{Dataset Overview}
\label{sec: Dataset Overview}

\begin{table*}[t]
\centering
\footnotesize
\setlength{\tabcolsep}{4.5pt}
\renewcommand{\arraystretch}{1}
\caption{Comparison with state-of-the-art arbitrary point tracking methods.
Training data: PO (Point Odyssey), Kub (Kubric), MF (MultiFlow), FE-FastKub (Our training dataset). speed expressed as ms per time step. Gray rows indicate our methods.}
\vspace{-0.2cm}
\label{tab:tap_results}
\resizebox{\textwidth}{!}{
\begin{tabular}{lcccccccccc}
\toprule
\multirow{2}{*}{Method} & \multirow{2}{*}{Train Data} & \multirow{2}{*}{Input} & \multirow{2}{*}{Time$\downarrow$} &
\multicolumn{3}{c}{InivTAP} & \multicolumn{3}{c}{DrivTAP} & \multirow{2}{*}{Mean $\delta_{\text{avg}}^{\text{vis}}$ $\uparrow$} \\
\cmidrule(lr){5-7} \cmidrule(lr){8-10}
& & & & AJ$\uparrow$ & $\delta_{\text{avg}}^{\text{vis}}\uparrow$ & OA$\uparrow$
& AJ$\uparrow$ & $\delta_{\text{avg}}^{\text{vis}}\uparrow$ & OA$\uparrow$ &  \\
\midrule
% ===================== Frame-based Methods =====================
PIPs++~\cite{zheng2023pointodyssey} & PO & Frame & 21.1 & -- & 31.2 & -- & -- & 37.8 & -- & 34.5 \\
Chrono~\cite{kim2025exploring} & Kub & Frame & -   & 31.2 & 44.7 & 79.8 & 15.8 & 30.8 & 68.1 & 37.8 \\
CoTracker3~\cite{karaev2025cotracker3} & Kub & Frame & 19.0 & 41.8 & 53.2 & 72.8 & \underline{37.1} & 46.5 & 95.4 & 49.9 \\
\rowcolor{gray!20} TAPFormer-F (Ours) & FastKub & Frame & \underline{16.5} & \underline{44.6} & 54.5 & 74.7 & 36.6 & \underline{46.7} & 93.8 & 50.6 \\
\midrule
% ===================== Event-based Methods =====================
E2Vid~\cite{rebecq2019events}-PIPs++~\cite{zheng2023pointodyssey} & PO & Event & 21.4 & -- & 14.3 & -- & -- & 30.4 & -- & 22.4 \\
E2Vid~\cite{rebecq2019events}-Chrono~\cite{kim2025exploring} & Kub & Event & - & 10.4 & 17.9 & 68.8 & 8.1 & 16.9 & 68.7 & 17.4 \\
E2Vid~\cite{rebecq2019events}-CoTracker3~\cite{karaev2025cotracker3} & Kub & Event & 22.2 & 12.8 & 19.4 & 56.6 & 18.6 & 27.0 & 89.4 & 23.2 \\
ETAP~\cite{hamann2025etap} & E-Kub & Event & 36.9 & 12.8 & 22.3 & \underline{86.3} & 13.5 & 27.8 & 68.1 & 25.1 \\
\rowcolor{gray!20} TAPFormer-E (Ours) & E-FastKub & Event & \textbf{14.0} & 20.9 & 29.8 & 79.0 & 28.9 & 37.6 & 90.2 & 33.7 \\
\midrule
% ===================== Frame-Event Fusion Methods =====================
FETAP~\cite{liu2025tracking} & MF & Frame+Event & - & -- & 25.7 & -- & -- & 30.3 & -- & 28.0 \\
FETAP~\cite{liu2025tracking} & FE-FastKub & Frame+Event & 23.4 & 42.2 & \underline{54.9} & 83.9 & 36.8 & 46.4 & 95.2 & \underline{50.7} \\
\rowcolor{gray!20} TAPFormer (Ours) & FE-FastKub & Frame+Event & 20.2 & \textbf{57.0} & \textbf{69.9} & \textbf{95.2} & \textbf{48.8} & \textbf{60.1} & \textbf{97.8} & \textbf{65.0} \\
\bottomrule
\end{tabular}
}
\vspace{-0.5cm}
\end{table*}

\textbf{Training Dataset Generation.} 
To train our model, we created a high-frame-rate synthetic dataset using the Kubric~\cite{greff2022kubric} simulation engine, named \textit{FE-FastKub}. Each scene was rendered at a resolution of 512$\times$512 and a frame rate of 48\,FPS over 2\,s, producing temporally dense trajectories that enable high-fidelity supervision.
Compared with the 12\,FPS MoVi dataset, this configuration provides finer temporal granularity and allows realistic event simulation.

For each sequence, both clean sharp frames and motion-blurred frames were rendered. The former were used to generate events, and the latter served as network inputs. The final dataset contains 10,953 samples, each with 96 RGB images, 1024 ground-truth trajectories at 48\,FPS, and the corresponding synthetic event stream generated by v2e~\cite{hu2021v2e}.
During training, we sample 12\,Hz image inputs while leveraging 48\,Hz trajectory supervision, allowing the model to learn the transient feature change between consecutive frames. Furthermore, 2,878 high-speed samples are included to increase motion blur and improve the model’s robustness to image degradation. This design helps the network rely less on high-quality RGB inputs and more on the temporal cues provided by events.

\textbf{Real Dataset Construction.}
% To evaluate real-world generalization, we constructed a novel frame-event TAP dataset. Two datasets were collected: 
To evaluate real-world generalization, we constructed a novel frame–event TAP dataset that comprises two subsets:
(i) InivTAP: sequences captured by a DAVIS346 camera (346$\times$260, 20\,FPS); and (ii) DrivTAP: sequences captured using our custom frame-event synchronization system, which integrates a Prophesee EVK4 (1280$\times$720, 120\,dB dynamic range) and an SG2-AR0231 RGB camera (1920$\times$1080, 22\,FPS).

The InivTAP subset contains 8 sequences of 5-7.5\,s each, averaging 12.8 annotated points per sequence at 20\,Hz. It covers a wide range of conditions, including fast motion, low light, overexposure, static scenes, dual-motion cases (both camera and object motion), and both indoor and outdoor environments.
The DrivTAP subset includes 5 driving sequences (5-10\,s each, average 12.2 points) captured at 10\,Hz with ground-truth supervision at 20\,Hz. 
Owing to the severe blur and overexposure, manually annotating a 5\,s sequence typically takes about two hours, with more detailed in the supplementary material.

In total, our real-world dataset provides 20,450 annotated points (10,850+9,600 across two subsets) from 13 diverse sequences, covering a variety of conditions. To our best knowledge, it represents the first real-world multimodal TAP benchmark designed for evaluating frame-event fusion methods under realistic and challenging settings.
\section{EXPERIMENTS}
\label{sec: EXPERIMENTS}

\subsection{Implementation Details}
\label{sec_:implementation}
Our model was trained on four NVIDIA RTX~4090 GPUs using images of 512$\times$512 resolution and a batch size of 1. We employed the AdamW optimizer with an initial learning rate of 5$\times$10$^{-4}$. Each sample contains 96 time steps and 1,024 ground-truth trajectories.
During training, 150 trajectories were randomly selected from each sample, forming a sequence of 24 consecutive steps. To encourage temporal consistency learning, trajectories that persist for longer durations were assigned a higher sampling probability.
All testing experiments were conducted on a laptop equipped with an NVIDIA RTX~3090 GPU.

\subsection{Task 1: TAP}
\label{sec_:tap}

\textbf{Evaluation Protocol.}
We evaluate our approach on our two challenging real-world datasets, \textit{InivTAP} and \textit{DrivTAP}.
\textit{InivTAP} covers complex representative challenging scenarios,
while \textit{DrivTAP} includes real driving sequences captured in both daytime and nighttime.
To assess long-term tracking stability and occlusion awareness, we follow the TAP-Vid~\cite{doersch2022tap} benchmark and report three metrics: the average visible-point duration under a pixel error threshold ($\delta^{\text{vis}}_{\text{avg}}$), Occlusion Accuracy (OA), and Average Jaccard (AJ).

We compare our method with three groups of baselines:
(i) sota frame-based trackers, including PIPs++~\cite{zheng2023pointodyssey}, CoTracker3~\cite{karaev2025cotracker3}, and Chrono~\cite{kim2025exploring};
(ii) event-based trackers, including ETAP~\cite{hamann2025etap}, as well as variants where events are reconstructed into videos via E2Vid~\cite{rebecq2019events} and then processed by frame-based methods; and
(iii) frame-event fusion trackers, represented by FETAP~\cite{liu2025tracking}.
Since FETAP does not explicitly model occlusions, we retrain it on our \textit{FE-FastKub} dataset to ensure a fair comparison.
All methods are carefully fine-tuned for each sequence. 
Additionally, we include single-modality variants of our network that process only frame or event inputs, without cross-modal fusion.

\textbf{Results on InivTAP.}
\begin{figure}[t]
   \centering
   \includegraphics[width=0.45\textwidth]{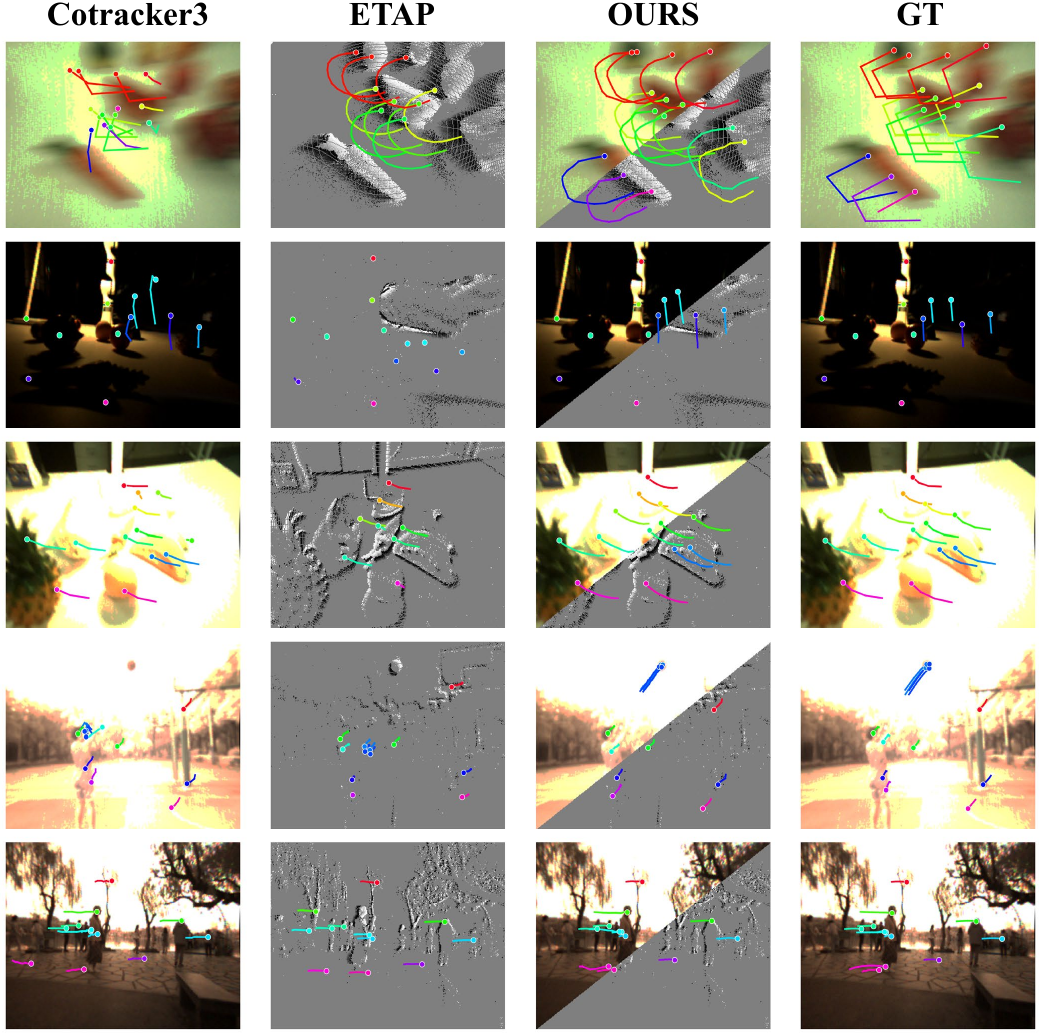} % 替换为你的图片文件名
   \vspace{-0.25cm}
   \caption{Task 1: TAP on \textit{InivTAP}. Rows show different sequences, and columns correspond to frame-based, event-based, fusion-based (ours), and ground-truth results.}
   \label{fig:inivdata_result}
   \vspace{-0.65cm}
\end{figure}
As shown in~\tabref{tab:tap_results}, event-based methods excel in fast motion and high-dynamic-range conditions due to their high temporal resolution, while frame-based methods achieve higher accuracy in static, well-lit scenes. 
However, the former can lack detailed spatial structure, and the latter degrades severely under motion blur or overexposure.
While naive RGB-event fusion mitigates these limitations, it fails to deal with their temporally-non-aligned nature. 
In contrast, our method adaptively integrates both cues, achieving consistent gains across all conditions. 
It attains accurate tracking under normal illumination by using detailed texture information from RGB frames, while leveraging events for rapid motion (see \figref{fig:inivdata_result}), quantitatively achieving a \textbf{36.4\%} AJ improvement over the frame-based CoTracker3 and \textbf{35.1\%} over the fusion-based FETAP. 
Our single-modality variants also outperform prior single-modality baselines, confirming the benefits of our high-frame-rate FastKubric training data and the proposed design of feature extraction.

\textbf{Results on DrivTAP.}
\begin{figure}[t]
   \centering
   \includegraphics[width=0.45\textwidth]{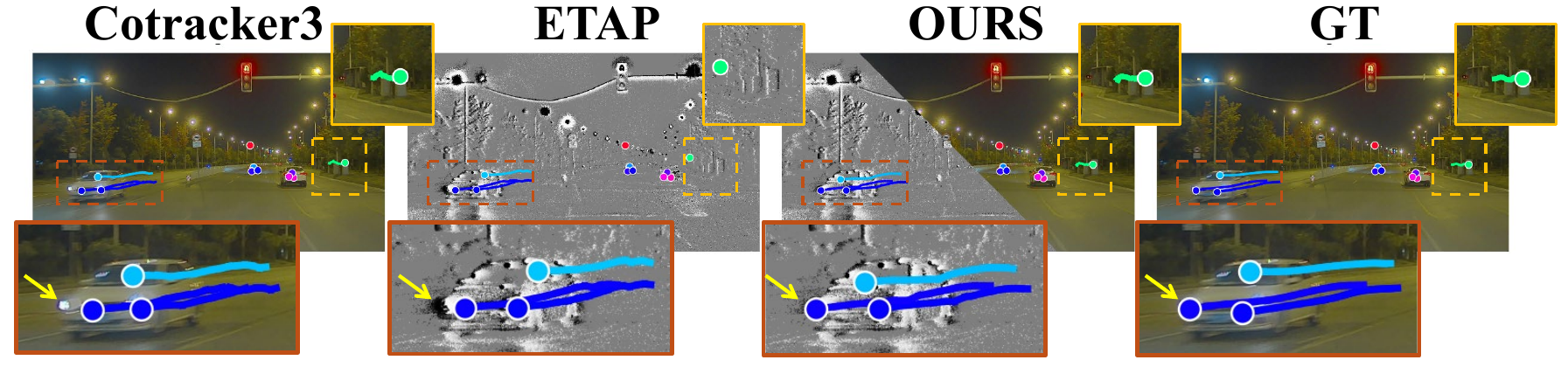} % 替换为你的图片文件名
   \vspace{-0.25cm}
   \caption{Red boxes show fast-moving vehicles where frame-based tracking drifts; yellow boxes highlight texture-similar regions where event-based tracking fails.
Our fusion-based method achieves stable and accurate tracking in both cases.}
   \label{fig:Driving_result}
   \vspace{-0.65cm}
\end{figure}
To emphasize the temporal resolution of event cameras, the ground-truth trajectories in \textit{DrivTAP} are annotated at twice the frame rate of the RGB images.
This dataset contains dynamic real-world driving scenarios with rapidly changing illumination, fast oncoming traffic, and vehicles moving at similar speeds to the ego-car, resulting in low relative motion and frequent motion blur, which making the TAP task extremely challenging.

As shown in ~\tabref{tab:tap_results}, the performance gap between frame-based and event-based methods becomes smaller in this dataset, since the frame rate of frame-based trackers is inherently limited by the RGB camera, while event-based methods benefit from their high temporal resolution.
Our proposed TAPFormer achieves the best overall performance, consistently surpassing both single-modality and previous fusion-based methods, see ~\figref{fig:Driving_result}.
Quantitatively, TAPFormer improves the AJ metric by \textbf{261.5\%} over the event-based ETAP, \textbf{31.5\%} over the frame-based CoTracker3, and \textbf{32.6\%} over the fusion-based FETAP.
Moreover, its inference speed remains comparable to mainstream trackers and even exceeds that of our single-modality baselines, demonstrating both efficiency and robustness in real-world driving scenarios.

\begin{table}[t]
\centering
\footnotesize
\setlength{\tabcolsep}{6pt}
\renewcommand{\arraystretch}{1}
\caption{Comparison of tracking accuracy on EDS and EC datasets.
Values are in percentages (\%). The best results are bold.}
\label{tab:eds_ec_comparison}
\vspace{-0.1cm}
\begin{tabular}{lcccccc}
\toprule
\multirow{2}{*}{Method} & \multirow{2}{*}{Input} &
\multicolumn{2}{c}{EDS} & \multicolumn{2}{c}{EC} \\
\cmidrule(lr){3-4} \cmidrule(lr){5-6}
 &  & FA$\uparrow$ & EFA$\uparrow$ & FA$\uparrow$ & EFA$\uparrow$ \\
\midrule
% ===================== Video-based =====================
PIPs++\cite{zheng2023pointodyssey} & Frame & 75.1 & 63.0 & 82.6 & 82.3 \\
CoTracker3\cite{karaev2025cotracker3} & Frame & 80.2 & 68.8 & 92.5 & 91.9 \\
Chrono\cite{kim2025exploring} & Frame & -- & -- & 82.9 & 82.3 \\
\rowcolor{gray!20} TAPFormer-F (Ours) & Frame & 80.9 & 69.1 & 92.7 & 92.1 \\
\midrule
% ===================== Event-based =====================
EM-ICP~\cite{conf_icra_ZhuAD17} & Event & 16.1 & 12.0 & 33.7 & 33.4 \\
HASTE~\cite{conf_bmvc_AlzugarayC20} & Event & 9.6 & 6.3 & 44.2 & 42.7 \\
AEB-Tracker~\cite{wang2024asynchronous} & Event & 47.3 & 46.6 & 55.3 & 47.3 \\
ETAP~\cite{hamann2025etap} & Event & 74.5 & 63.9 & 88.1 & 87.6 \\
MATE~\cite{han2025mate} & Event & -- & -- & 88.5 & 87.5 \\
\rowcolor{gray!20} TAPFormer-E (Ours) & Event & 76.8 & 64.5 & 86.6 & 86.1 \\
\midrule
% ===================== Fusion-based =====================
EKLT~\cite{journals_ijcv_GehrigRGS20} & F + E & 32.5 & 20.5 & 81.1 & 77.5 \\
DeepEvT~\cite{conf_cvpr_MessikommerFG023} & F + E & 61.3 & 50.5 & 82.5 & 81.8 \\
FETAP~\cite{liu2025tracking} & F + E & 72.2 & 63.2 & 84.4 & 83.8 \\
\rowcolor{gray!20} TAPFormer (Ours) & F + E & \textbf{82.3} & \textbf{70.4} & \textbf{93.3} & \textbf{92.6} \\
\bottomrule
\end{tabular}
\vspace{-0.2cm}
\end{table}

\subsection{Task 2: Feature Tracking}
\label{sec_:feature track}
\begin{figure}[t]
   \centering
   \includegraphics[width=0.45\textwidth]{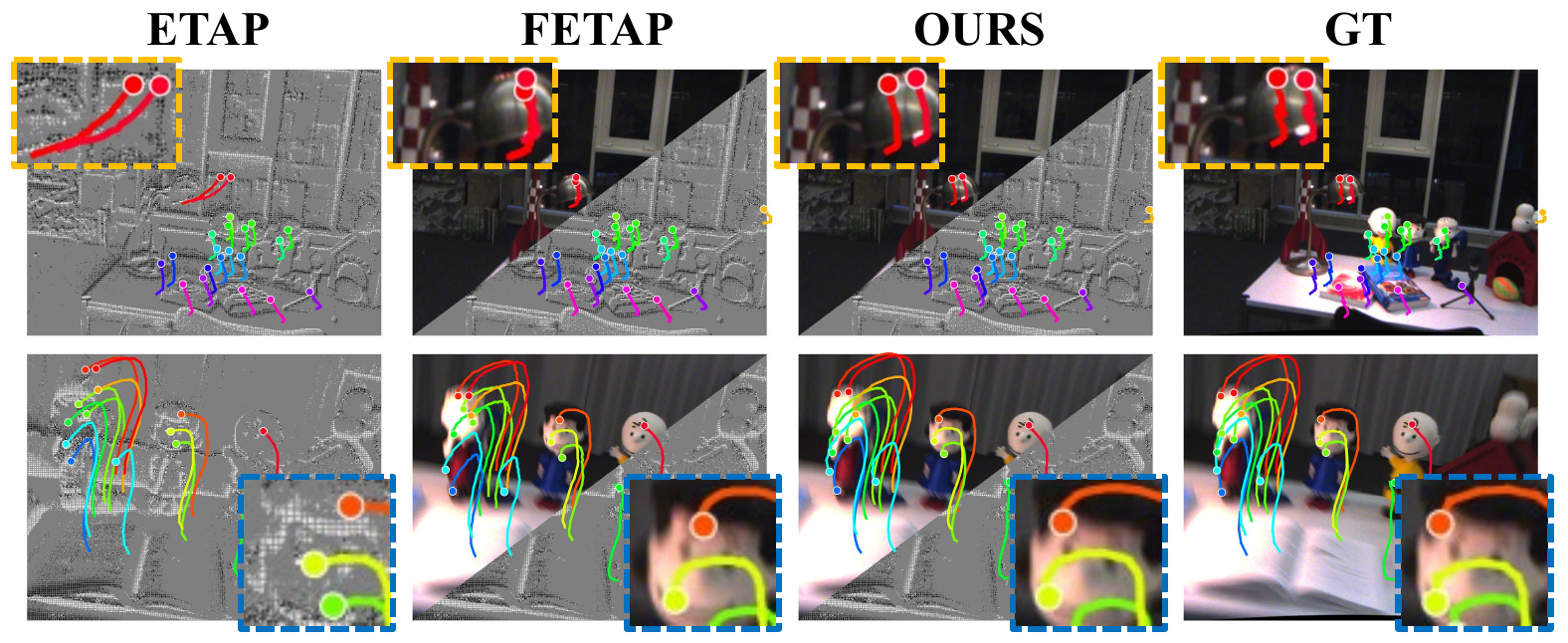} % 替换为你的图片文件名
   \vspace{-0.2cm}
   \caption{Task 2: Feature tracking on EDS. }
   \label{fig:EDS_result}
   \vspace{-0.65cm}
\end{figure}

We further evaluate our method on the \textit{EC}~\cite{mueggler2017event} and \textit{EDS}~\cite{hidalgo2022event} datasets for feature point tracking. Both datasets provide synchronized RGB and event streams at resolutions of 240$\times$180 and 640$\times$480, respectively. Following standard practice~\cite{conf_cvpr_MessikommerFG023}, we report Feature Age (FA) and Expected Feature Age (EFA), which measure the temporal consistency of tracked points under a fixed tracking error threshold.
We observed that the \textit{Peanuts Light} sequence in the EDS dataset contains several outlier trajectories that noticeably distort evaluation metrics. So, we apply a simple temporal smoothing to the ground-truth trajectories of this sequence (see supplementary material).

We compare our approach with state-of-the-art frame-only, event-only, and frame-event fusion trackers (~\tabref{tab:eds_ec_comparison}). Frame-based TAP methods (PIPs++~\cite{zheng2023pointodyssey}, CoTracker3~\cite{karaev2025cotracker3}, and Chrono~\cite{kim2025exploring}) perform well in static scenes but degrade under motion blur and low resolution. Our frame-only baseline achieves higher accuracy on both datasets, demonstrating the effectiveness of our feature extraction design. For event-based tracking, we include EM-ICP~\cite{conf_icra_ZhuAD17}, HASTE~\cite{conf_bmvc_AlzugarayC20}, AEB-Tracker~\cite{wang2024asynchronous}, ETAP~\cite{hamann2025etap}, and MATE~\cite{han2025mate}. These methods benefit from the high temporal resolution of event data but lack semantic and spatial cues, which limit their robustness under complex textures. Fusion-based methods such as EKLT~\cite{journals_ijcv_GehrigRGS20}, DeepEvT~\cite{conf_cvpr_MessikommerFG023}, and FETAP~\cite{liu2025tracking} combine both modalities but do not fully exploit their complementary strengths, leading to suboptimal temporal stability.

Across both datasets, TAPFormer achieves the best overall performance. It consistently improves EFA compared with all baselines, maintaining smooth and stable trajectories even under fast motion and partial occlusions, as shown in ~\figref{fig:EDS_result}. These results confirm that our method effectively integrates the complementary advantages of frames and events, achieving temporally consistent and fine-grained tracking in complex real-world scenes.

\subsection{Ablation Study}
\label{sec_:ablation}
\begin{table}[t]
\centering
\footnotesize
\setlength{\tabcolsep}{5pt}
\renewcommand{\arraystretch}{1}
\caption{Ablation study on the EDS dataset.
FA and EFA denote feature age and expected feature age, respectively. MSSF and TAM denote multi-scale semantic features and temporal attention module, respectively.}
\vspace{-0.15cm}
\label{tab:ablation}
\begin{tabular}{ccccc|cc}
\toprule
\textbf{FE-FastKub} & \textbf{CLWF} & \textbf{TAF} & \textbf{MSSF} & \textbf{TAM} & \textbf{FA$\uparrow$} & \textbf{EFA$\uparrow$} \\
\midrule
 &  &  & & & 0.646 & 0.535 \\
\checkmark & &  &  &  & 0.701 & 0.585 \\
\checkmark & \checkmark &  &  &  & 0.763 & 0.647 \\
\checkmark & \checkmark & \checkmark &  &  & 0.803 & 0.685 \\
\checkmark & \checkmark & \checkmark & \checkmark &  & 0.814 & 0.698 \\
\checkmark & \checkmark & \checkmark &  & \checkmark & 0.807 & 0.691 \\
\checkmark & \checkmark & \checkmark & \checkmark & \checkmark & \textbf{0.823} & \textbf{0.704} \\
\bottomrule
\end{tabular}
\vspace{-0.6cm}
\end{table}

\textbf{Component Analysis.}
We conduct a detailed ablation study on the \textit{EDS} dataset to assess the contribution of each proposed component, including the training dataset \textit{FE-FastKub} dataset, TAF mechanism, CLWF module, the multi-scale semantic features (MSSF), and the temporal attention module (TAM).
As summarized in ~\tabref{tab:ablation}, the baseline model was trained on the MultiFlow dataset without occlusion annotations. When the fusion module is removed, frame-event fusion is replaced by a channel-wise concatenation followed by a convolutional layer.

The results show that both the dataset and fusion strategy are crucial for high-quality tracking.
Introducing the proposed fusion module yields a notable gain in FA and EFA, validating the effectiveness of transient asynchronous feature aggregation.
Incorporating multi-scale semantic representations enhances spatial discriminability, while the temporal attention module further improves feature consistency across time.
The complete model achieves the highest FA (0.823) and EFA (0.704), demonstrating that each component contributes progressively to the overall performance.

\begin{figure}[t]
   \centering
   \includegraphics[width=0.45\textwidth]{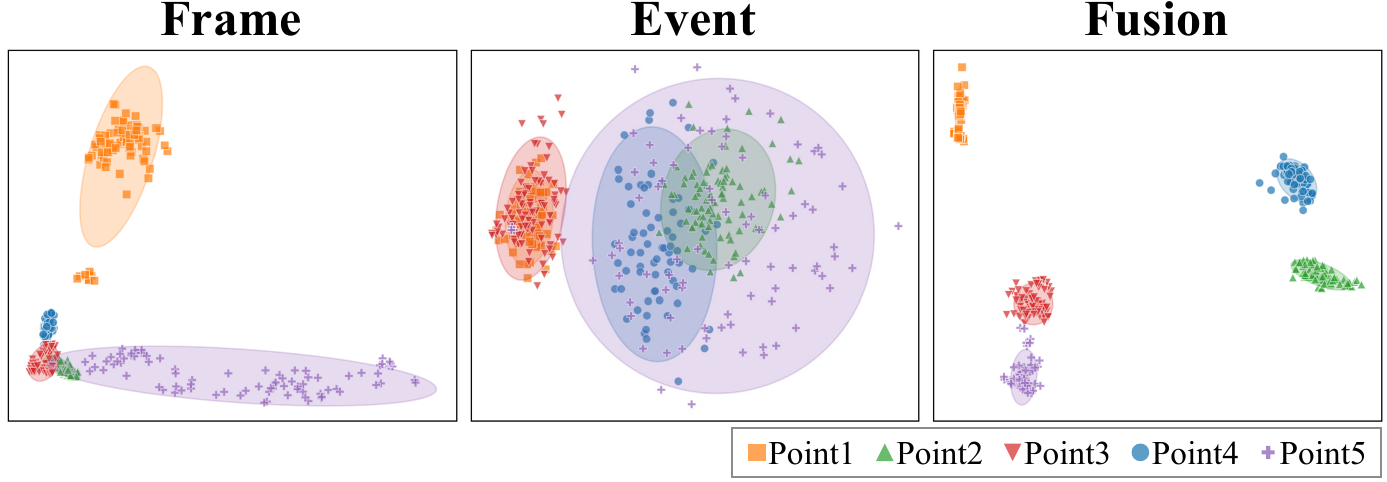} % 替换为你的图片文件名
   \vspace{-0.2cm}
   \caption{PCA visualization of feature embeddings from frame-, event-, and fusion-based representations. Each color denotes a tracked point; Ellipses indicate the covariance-based dispersion of feature embeddings in the PCA space, reflecting the intra-point stability and inter-point separability of learned representations.}
   \label{fig:feature_vector_clust}
   \vspace{-0.65cm}
\end{figure}

\textbf{Single-Modality Evaluation.} We perform single-modality evaluations in~\tabref{tab:tap_results} and ~\tabref{tab:eds_ec_comparison}.
Our frame-only and event-only variants consistently outperform previous single-modality trackers, confirming the benefits of the proposed training data and feature extraction backbone.
Nevertheless, each modality exhibits inherent weaknesses: the frame-only tracker degrades under fast motion or overexposure, whereas the event-only tracker lacks spatial texture cues.
By jointly leveraging both modalities, our fusion-based design effectively mitigates these limitations and maintains robust tracking across diverse scenarios.

To further examine the robustness of the learned features and to validate the effectiveness of our cross-modal local weighted fusion module, we compare the temporal dynamics of point feature vectors obtained from the frame-only, event-only, and fused models. For each method, the same tracking points are sampled along ground-truth trajectories, and their feature vectors are projected onto a 2D space via principal component analysis (PCA). As shown in~\figref{fig:feature_vector_clust}, temporally consistent and robust features are expected to form compact clusters for the same point while maintaining clear separations across different points. The fused representations exhibit both tight intra-point clustering and large inter-point margins over time, indicating superior temporal coherence and discriminative embedding quality than single-modality baselines.

\begin{figure}[t]
   \centering
   \includegraphics[width=0.45\textwidth]{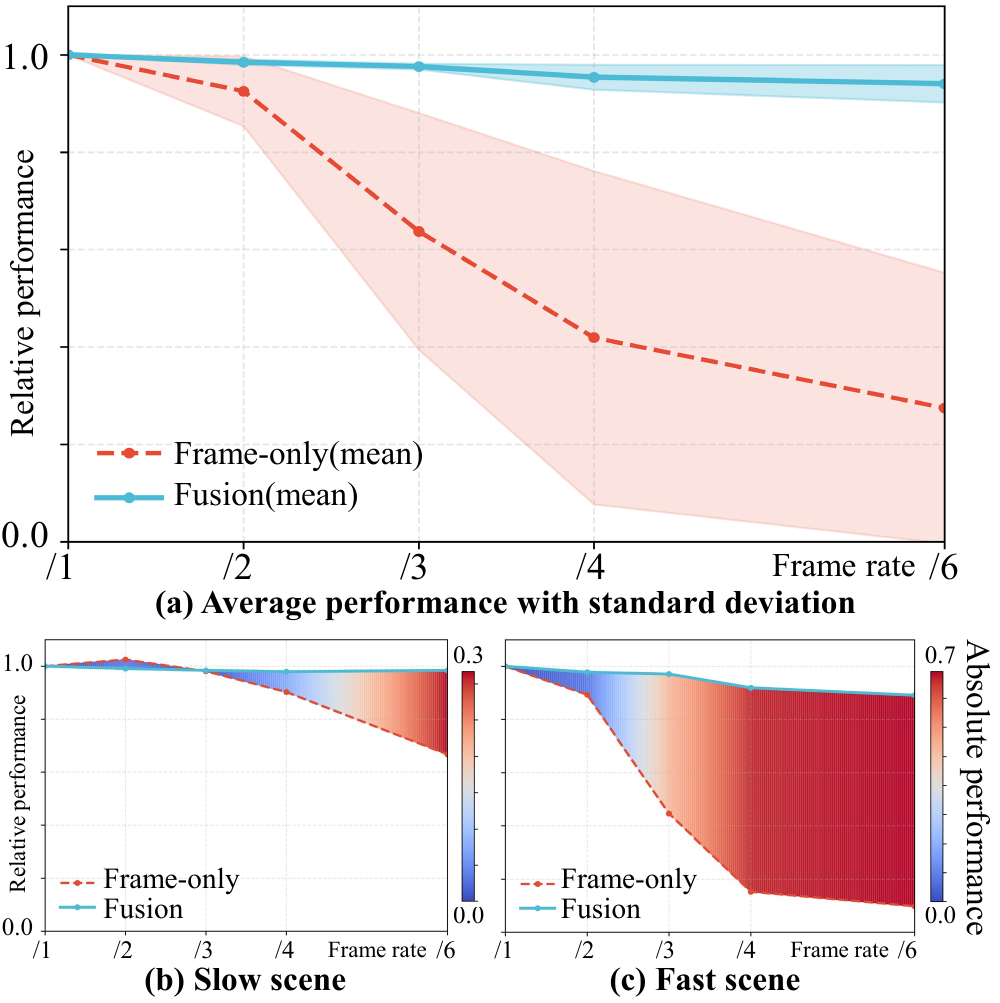} % 替换为你的图片文件名
   \vspace{-0.1cm}
   \caption{Frame-rate sensitivity comparison between CoTracker3 and our method.
(a) shows overall performance trends with variance shading; (b–c) visualize relative performance in slow and fast motion scenarios, where color bands denote absolute performance differences (see color bar on the right).}
   \label{fig:framerate_compare}
   \vspace{-0.5cm}
\end{figure}

\textbf{Frame Rate Sensitivity.} 
We further analyzed how the input frame rate affects tracking performance FA for our method and CoTracker3 under slow, normal, and fast motion settings. see ~\tabref{tab:fps}
As the frame rate decreases, CoTracker3 suffers severe degradation and becomes nearly invalid at 10\, Hz.
In contrast, our fusion-based tracker remains stable, exhibiting only a 6.5\% reduction in performance compared to a 75.3\% drop for the frame-only method, as shown in ~\figref{fig:framerate_compare} 
This experiment highlights the adaptability of our transient asynchronous fusion mechanism to low-frame-rate and high-speed conditions, demonstrating its potential for real-world deployment where frame capture rates are constrained.

\begin{table}[t]
\centering
\footnotesize
\setlength{\tabcolsep}{4.5pt}
\renewcommand{\arraystretch}{1}
\caption{Effect of frame rate on tracking performance.
Results are reported on the EDS dataset using different input frame rates.}
\vspace{-0.1cm}
\label{tab:fps}
\begin{tabular}{l|cccccc}
\toprule
\multirow{2}{*}{\textbf{Method}} & \multicolumn{6}{c}{\textbf{Frame rate (FPS)}} \\
\cmidrule(lr){2-7}
 & 75 & 37.5 & 25 & 18.75 & 12.5 & 9.375 \\
\midrule
CoTracker3 & 83.9 & 74.9 & 51.2 & 32.4 & 20.7 & 13.2 \\
Ours & \textbf{85.1} & \textbf{83.2} & \textbf{82.7} & \textbf{80.8} & \textbf{79.6} & \textbf{75.8} \\
\bottomrule
\end{tabular}
\vspace{-0.2cm}
\end{table}

\section{Conclusion}

We introduced TAPFormer, a synergistic frame-event tracker for robust and high-frequency point tracking under diverse visual conditions.
Our framework builds upon two core innovations: a transient feature update mechanism that preserves the fine temporal dynamics of event streams, and a cross-modal local weighted fusion module that adaptively integrates complementary cues from frames and events to produce stable and discriminative representations.
Extensive evaluations show that TAPFormer consistently achieves state-of-the-art performance on TAP and feature tracking benchmarks, outperforming all existing frame-, event-, and fusion-based baselines.

Our work also contributes two key resources to advance frame-event fusion research: a large-scale synthetic dataset for high-frame-rate supervision, and the first real-world, manually annotated TAP dataset with synchronized frame-event sequences under challenging illumination and motion. 
We hope these resources and the proposed framework will provide a foundation for advancing robust visual tracking in complex environments.

\clearpage
\setcounter{page}{1}
\maketitlesupplementary

\section{Additional Method Details}
\paragraph{Transformer-based Trajectory Refinement.}
In the main paper, we introduced the multimodal fusion network and the construction of the multi-scale fused feature pyramid. Here, we describe how these features are used in a transformer-based trajectory optimizer (based on Cotracker3\cite{karaev2025cotracker3}) to iteratively refine the point coordinates and visibility. 

Let \(\mathcal{P}=\{\mathbf{P}^{(l)}\,|\, l=0,1,2\} \)
denote the three-level fused feature pyramid extracted from our model.  
Given a temporal window of size $W=16$, the initial point state is representing the coordinate sequence and visibility estimates:
\begin{equation}
\mathbf{x}=\{(x_t,y_t)\}_{t=1}^{W}, \qquad 
\mathbf{v}=\{v_t\}_{t=1}^{W},
\end{equation}

At each refinement iteration, we extract a local feature patch around the predicted coordinates from each pyramid level. For scale $l$, a $(2r+1)\times(2r+1)$ patch with $r=3$ is sampled using an operator $S(\cdot)$:
\begin{equation}
\mathbf{f}^{(l)}_t = S\!\left(\mathbf{P}^{(l)}_t,\,(x_t,y_t),\,r\right)
\in \mathbb{R}^{(2r+1)\times(2r+1)\times C_l}.
\end{equation}

To measure temporal consistency across the window, we compute correlations between 
$\mathbf{f}^{(l)}_t$ and the corresponding sampled patches at all other frames, resulting in a correlation matrix of size $(2r+1)^2 \times (2r+1)^2$.  
Each matrix is flattened and encoded by an MLP, and embeddings from all three scales are concatenated to obtain the final correlation descriptor.

We further include a positional encoding of relative motion across frames.  
The correlation embeddings, visibility estimates, and motion encoding are concatenated and passed through a spatio-temporal transformer, which predicts residual updates:
\begin{equation}
\mathbf{x} \leftarrow \mathbf{x} + \Delta\mathbf{x}, \qquad
\mathbf{v} \leftarrow \mathbf{v} + \Delta\mathbf{v}.
\end{equation}

We perform three refinement iterations.  
Thanks to the discriminative and temporally consistent fused feature pyramid, our optimizer converges quickly, unlike prior approaches (e.g., CoTracker3\cite{karaev2025cotracker3}) that typically require six iterations.  
This allows us to maintain high accuracy while significantly reducing computational cost.

\section{Additional Ablation Experiments}
\paragraph{Event Representation.}
\begin{table}[t]
\centering
\footnotesize
\setlength{\tabcolsep}{4.5pt}
\renewcommand{\arraystretch}{1}
\caption{Effect of event representation on tracking performance.}
\vspace{-0.2cm}
\label{tab:event_representation}
\begin{tabular}{l|cccc}
\toprule
\multirow{2}{*}{Representation} & \multicolumn{2}{c}{EDS} & \multicolumn{2}{c}{EC}\\
\cmidrule(lr){2-3} \cmidrule(lr){4-5}
 & FA$\uparrow$ & EFA$\uparrow$ & FA$\uparrow$ & EFA$\uparrow$ \\
\midrule
Event image\cite{maqueda2018event} & 81.4 & 69.5 & 90.8 & 90.2 \\
Voxel grid\cite{zhu2019unsupervised} & 82.5 & 69.8 & 92.1 & 91.5 \\
Time surface\cite{lagorce2016hots} & 82.3 & 70.4 & 93.3 & 92.6 \\
\bottomrule
\end{tabular}
\vspace{-0.2cm}
\end{table}

\begin{figure}[t]
   \centering
   \includegraphics[width=0.45\textwidth]{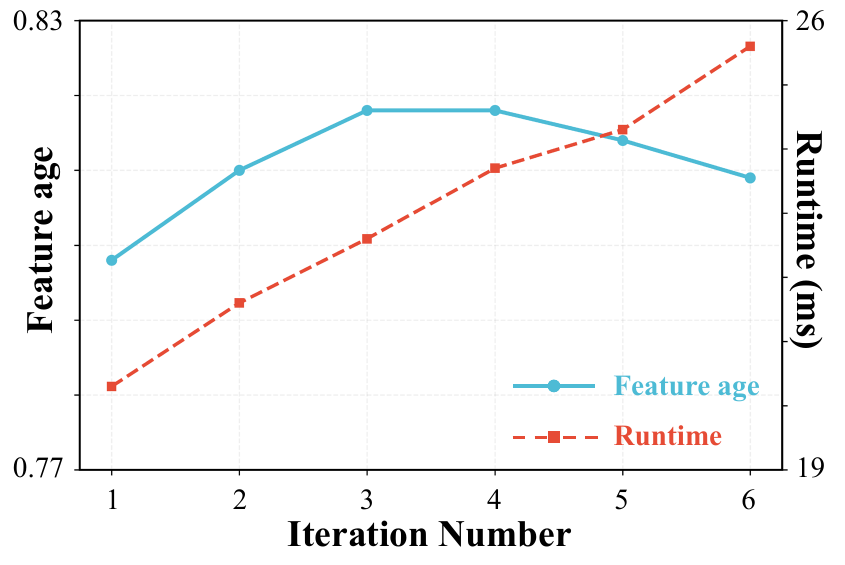} % 替换为你的图片文件名
   \vspace{-0.3cm}
   \caption{Performance and efficiency across tracking iterations. The left y-axis reports the feature age (higher is better), while the right y-axis shows the inference throughput measured in frames per millisecond (FPS/ms). The x-axis denotes the number of trajectory refinement iterations. This dual-axis plot illustrates how tracking accuracy and computational efficiency evolve as the number of iterations increases. }
   \vspace{-0.5cm}
   \label{fig:iteration}
\end{figure}

% To utilize asynchronous event streams containing rich spatio-temporal information as input to neural networks, the discrete events are typically converted into dense tensor-like representations. This conversion offers two main advantages: (1) it enables compatibility with standard convolutional architectures, and (2) by aggregating event data over a short time interval, it mitigates the impact of individual noisy events and improves robustness.

% We compare three widely used event representations: Event Image\cite{maqueda2018event}, Voxel Grid\cite{zhu2019unsupervised}, and Time Surfaces\cite{lagorce2016hots}. Following the Stacking Based on Time (SBT) scheme\cite{wang2019event}, we divide a fixed temporal window into five equal sub-windows. For each sub-window, the three representations are computed independently. The Event Image accumulates event counts, the Voxel Grid forms a discretized spatio-temporal volume, while Time Surfaces store the most recent event timestamp in a decay-like form. We test all representations on the EDS and EC dataset using identical training settings, and the quantitative results are reported in \tabref{tab:event_representation}. Time Surfaces achieve the best performance among the three.
To leverage the spatio-temporal information in asynchronous event streams, the sparse events are typically converted into dense tensor-like representations. This conversion (i) enables direct use of standard convolutional architectures and (ii) improves robustness by aggregating events within short temporal intervals.

We evaluate three widely used representations: Event Image~\cite{maqueda2018event}, Voxel Grid~\cite{zhu2019unsupervised}, and Time Surfaces~\cite{lagorce2016hots}. Following the Stacking Based on Time (SBT) scheme~\cite{wang2019event}, a fixed temporal window is divided into five sub-windows, and each representation is computed separately for every sub-window. Event Image accumulates event counts, Voxel Grid constructs a discretized spatio-temporal volume, and Time Surfaces encode the latest event timestamp in a decay-like form. All representations are evaluated on the EDS and EC datasets under identical training settings, and the quantitative results feature age (FA) and experted feature age (EFA), which quantify the duration until a track deviates beyond a threshold distance from the GT, are summarized in \tabref{tab:event_representation}. Given its higher accuracy and lower computational cost, we adopt Time Surfaces as our event representation.

\paragraph{Iteration Number.}
We analyze how the number of refinement iterations influences both accuracy and runtime. Increasing iterations improves performance but introduces an almost linear increase in computation. As shown in the two-axis performance–time curve (see~\figref{fig:iteration}), performance saturates around the third iteration: subsequent iterations yield only marginal gains while noticeably increasing cost. Thus, we adopt three iterations to balance accuracy and efficiency. This fast convergence reflects the stability and discriminative strength of our fused features, which provide reliable cues for trajectory refinement.

\section{Extended Experimental Results}
\begin{figure}[t]
   \centering
   \includegraphics[width=0.45\textwidth]{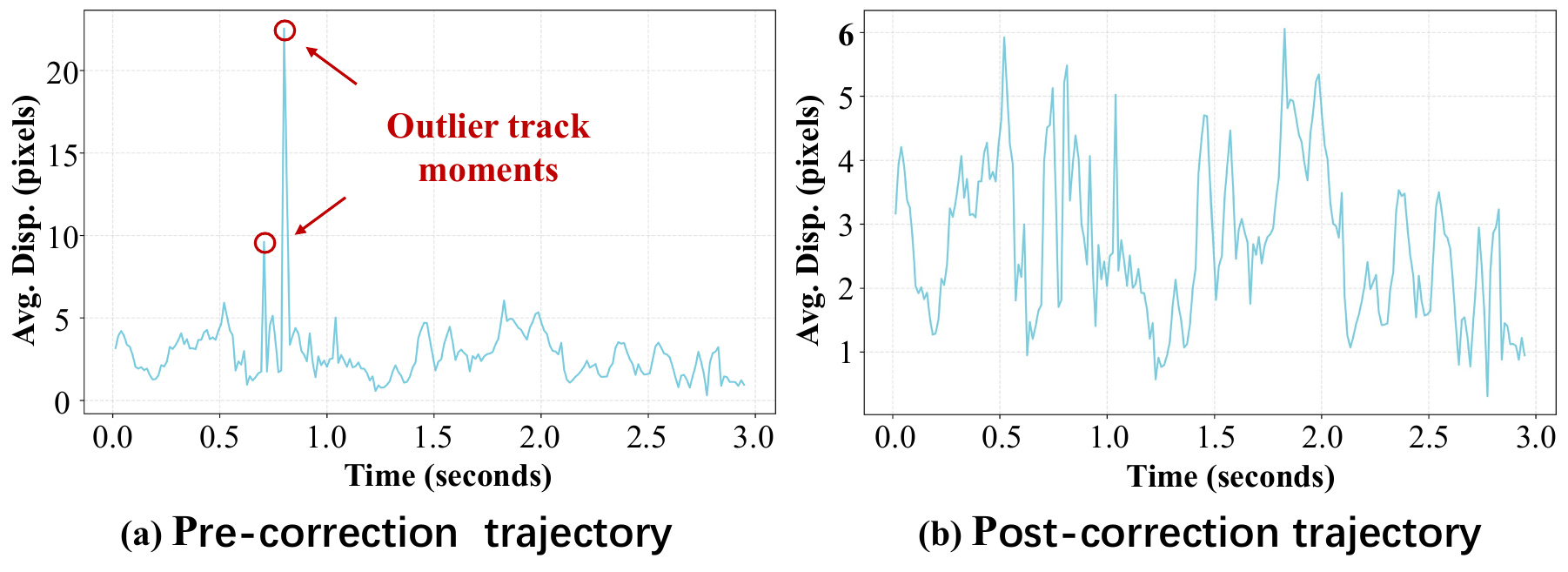} 
   \vspace{-0.3cm}
   \caption{Ground-truth trajectory correction for the peanuts light sequence. (a) Pre-correction trajectory with two displacement outliers. (b) Post-correction trajectory after smoothing, producing a stable and reliable ground truth. }
   \vspace{-0.5cm}
   \label{fig:gt_correct}
\end{figure}

\begin{figure}[t]
   \centering
   \includegraphics[width=0.45\textwidth]{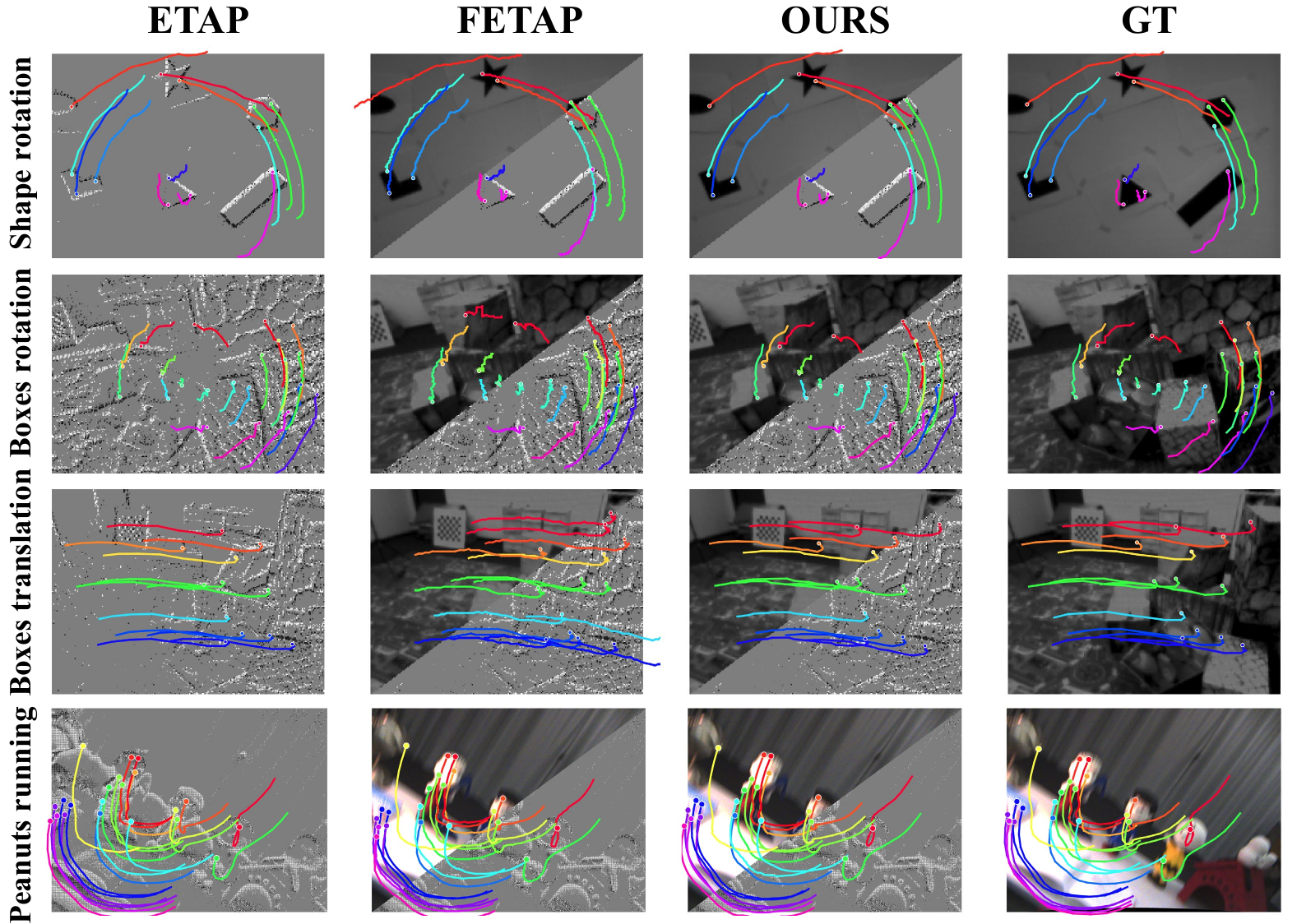} % 替换为你的图片文件名
   \vspace{-0.1cm}
   \caption{Additional visualizations on EC and EDS dataset. }
   \vspace{-0.65cm}
   \label{fig:ec_result}
\end{figure}
We provide additional quantitative and qualitative results to complement the main paper. ~\tabref{tab:scenario} reports the detailed results on the InivTAP dataset. ~\tabref{tab:eds_full_result} and ~\tabref{tab:ec_full_result} list the full evaluation results on the EDS and EC datasets. ~\figref{fig:ec_result} presents extra qualitative comparisons.

We additionally perform ground-truth trajectory correction for the peanuts light sequence in the EDS dataset. As shown in ~\figref{fig:gt_correct}, the original ground-truth trajectory contains two displacement outlier points, which we smooth during correction. As reported in ~\tabref{tab:eds_full_result}, the correction leads to performance improvements across all sequences and yields more distinguishable evaluation metrics.
\begin{table}[t]
\centering
\footnotesize
\setlength{\tabcolsep}{6pt}
\renewcommand{\arraystretch}{1.1}
\caption{Performance comparison across different scenarios and modalities. }
\label{tab:scenario}
\vspace{-0.2cm}
\begin{tabular}{llcccc}
\toprule
Scenario & Method & Modality & AJ$\uparrow$ &
$\delta_{\text{avg}}^{\text{vis}}$ $\uparrow$ & OA$\uparrow$ \\
\midrule
\multirow{3}{*}{Normal} 
 & ETAP       & Event           & 19.7 & 33.6 & 88.8 \\
 & CoTracker3 & Frame           & 61.3 & 77.1 & 92.1 \\
 & Ours  & F + E           & 71.1 & 84.8 & 92.7 \\
\midrule
\multirow{3}{*}{Fast} 
 & ETAP       & Event           & 27.6 & 40.6 & 89.2 \\
 & CoTracker3 & Frame           & 28.0 & 36.7 & 73.5 \\
 & Ours  & F + E           & 39.5 & 52.3 & 95.4 \\
\midrule
\multirow{3}{*}{Overexposure} 
 & ETAP       & Event           & 11.8 & 22.3 & 77.8 \\
 & CoTracker3 & Frame           & 30.2 & 43.9 & 70.6 \\
 & Ours  & F + E           & 38.6 & 50.7 & 89.6 \\
\midrule
\multirow{3}{*}{Static} 
 & ETAP       & Event           & 10.6 & 22.3 & 71.8 \\
 & CoTracker3 & Frame           & 53.4 & 72.2 & 88.6 \\
 & Ours  & F + E           & 78.9 & 90.0 & 94.9 \\
\bottomrule
\end{tabular}
\vspace{-0.5cm}
\end{table}

\section{Application to SLAM}
% To further demonstrate the practical value of our tracking algorithm, we integrate TAPFormer into a visual SLAM pipeline and evaluate its performance under challenging scenarios. Our system consists of three components: feature detection, feature tracking, and backend pose optimization. We adopt SuperPoint\cite{detone2018superpoint} for keypoint extraction and ensure well-distributed features. Then we apply a lightweight keypoint management module to maintain stable coverage. The feature tracks are produced using TAPFormer, providing robust and accurate correspondences across frames. For the backend, we reuse the optimization module of VINS-Mono\cite{qin2018vins} to estimate high-frequency camera poses.

To demonstrate the practical value of our tracker, we integrate TAPFormer into a visual SLAM pipeline and evaluate it under challenging scenarios. The system includes feature detection, feature tracking, and backend pose optimization. We use SuperPoint \cite{detone2018superpoint} to extract well-distributed keypoints, followed by a lightweight management module to maintain stable feature coverage. TAPFormer then produces robust inter-frame correspondences. For the backend, we directly reuse the VINS-Mono \cite{qin2018vins} optimization module to estimate high-frequency camera poses.

\paragraph{Evaluation Metric and Baselines.}
To evaluate velocity and pose accuracy in a principled manner, we adopt the speed-weighted success metric commonly used in recent event-based SLAM work. Let $\mathrm{RVE}_i$ denote the relative velocity error at timestamp $i$, $\mathbf{v}_{\mathrm{gt},i}$ the ground-truth velocity, and $\xi$ an error threshold. The success score is defined as
\begin{equation}
S_{\xi} = 
\frac{\sum_{i=1}^{N} \|\mathbf{v}_{\mathrm{gt},i}\| \cdot 
\delta(\mathrm{RVE}_i < \xi)}
{\sum_{i=1}^{N} \|\mathbf{v}_{\mathrm{gt},i}\|},
\end{equation}
where $\delta(\cdot)$ is the Dirac indicator.
This formulation emphasizes accurate estimates during high-speed motion by weighting each success according to the magnitude of the ground-truth velocity. By sweeping $\xi$ over the interval $[0,1]$, a success-rate curve is obtained, and the final AUC$_v$  provides a unified, speed-aware scalar metric. 
% This addresses the limitations of traditional metrics such as RPE and RVE, which do not sufficiently discriminate errors occurring at high velocities or provide a consistent ranking.
We compare our system against two event-based SLAM baselines: 1) SuperEvent\cite{burkhardt2025superevent} + OKVIS2\cite{leutenegger2022okvis2}, which uses stereo events, stereo images, and IMU measurements; 2) SDEVO\cite{zhong2025deep}, which operates purely on stereo events.

\begin{table}[t]
\centering
\footnotesize
\setlength{\tabcolsep}{5pt}
\renewcommand{\arraystretch}{1}
\caption{Comparison with event-based SLAM methods. We compare our approach with two event-based SLAM baselines. After directly replacing the VINS-Mono frontend with ours, our method achieves the best performance.}
\label{tab:slam_comparison}
\begin{tabular}{lcccc}
\toprule
\multirow{2}{*}{Method} &
\multicolumn{3}{c}{Modalities} & \multirow{2}{*}{AUC$_v$$\uparrow$} \\
\cmidrule(lr){2-4}
 & Imu Used & Events Used & Images Used & \\
\midrule
% ===================== Video-based =====================
SDEVO~\cite{zhong2025deep} & none & stereo & none & 6.13 \\
SuperEvent~\cite{burkhardt2025superevent} & yes & stereo & stereo & 6.27 \\
Ours & yes & monocular & monocular & \textbf{6.29} \\
\bottomrule
\end{tabular}
\vspace{-0.3cm}
\end{table}

\begin{figure}[t]
   \centering
   \includegraphics[width=0.45\textwidth]{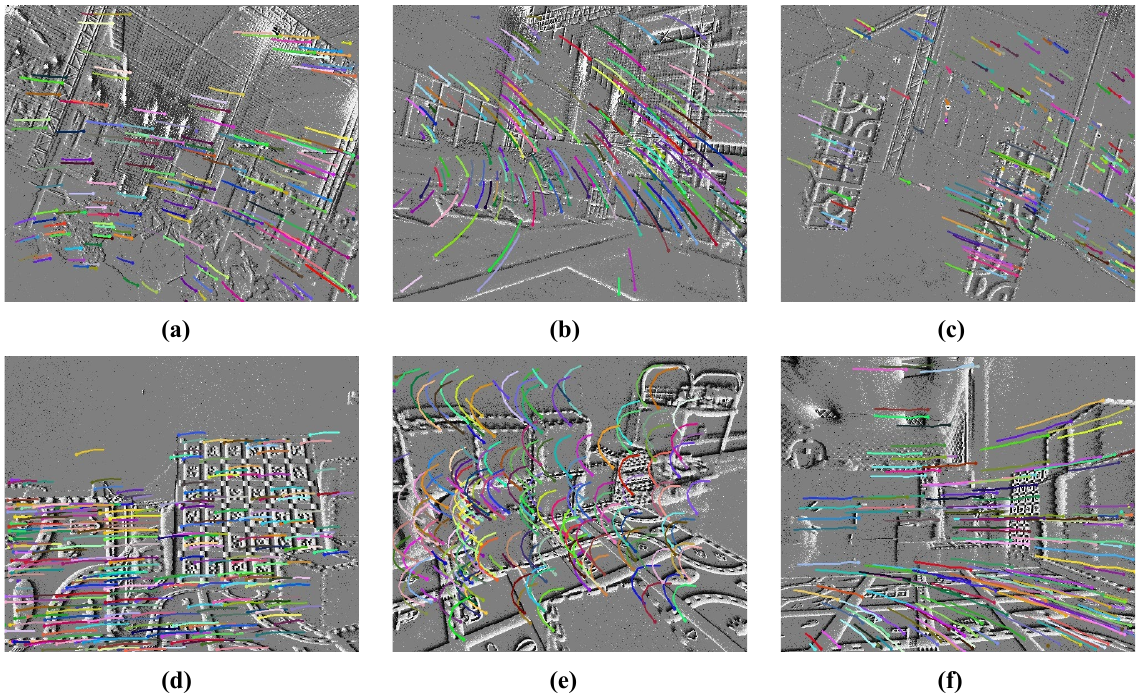} 
   \vspace{-0.2cm}
   \caption{Visualization of our tracking results on SLAM sequences. (a–c) show results on high-speed UAV scenarios, while (d–e) present tracking results on ground-based high-speed sequences. }
   \vspace{-0.2cm}
   \label{fig:slam_result}
\end{figure}

\paragraph{Results.}
~\figref{fig:slam_result} presents qualitative tracking results by TAPFormer, and ~\tabref{tab:slam_comparison} summarizes the final AUC$_v$ scores. Our approach achieves the best performance among all evaluated methods. 
% We attribute this improvement to two key reasons. First, TAPFormer enables arbitrary-frame-rate point tracking, allowing the backend to estimate poses at any desired timestamp without relying on linear interpolation. Second, by effectively combining the complementary strengths of frames and events, the system maintains high pose accuracy across both high-speed and low-speed motions. Together, these factors lead to consistently superior trajectory and velocity estimation across all tested sequences.

\section{Dataset Details}

We construct a large-scale synthetic dataset and provide two real, manually annotated frame-event TAP test sets. In the following, we describe the datasets in detail.

\paragraph{FE-FastKub}
Our FE-FastKub dataset differs significantly from existing TAP training sets. Specifically, it not only contains high-frame-rate photorealistic RGB images (with motion blur considered) but also provides corresponding synthetic event streams. To enhance the model's adaptability to fast-moving scenes, we further include explicit modeling of rapid motion scenarios. See ~\tabref{tab:datasets}

\begin{figure}[t]
   \centering
   \includegraphics[width=0.45\textwidth]{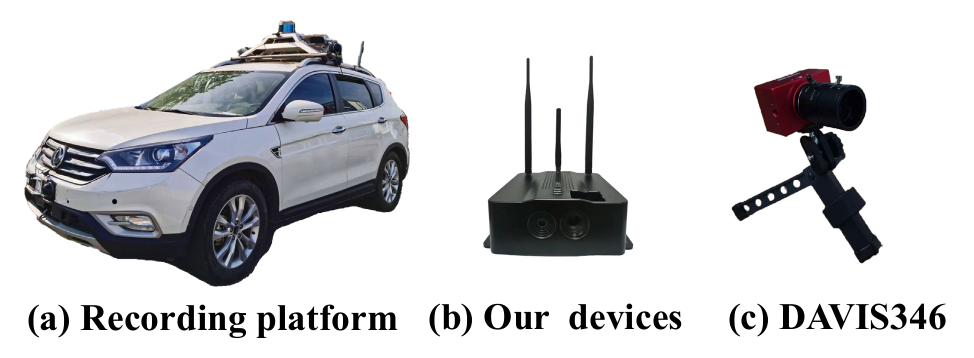} % 替换为你的图片文件名
   \vspace{-0.3cm}
   \caption{Data collection setup. (b) Our custom synchronized capture device is mounted on the (a) vehicle platform to record DrivTAP, and (c) a DAVIS346 camera is used to collect the InivTAP dataset. }
   \vspace{-0.1cm}
   \label{fig:device}
\end{figure}

\begin{figure}[t]
   \centering
   \includegraphics[width=0.45\textwidth]{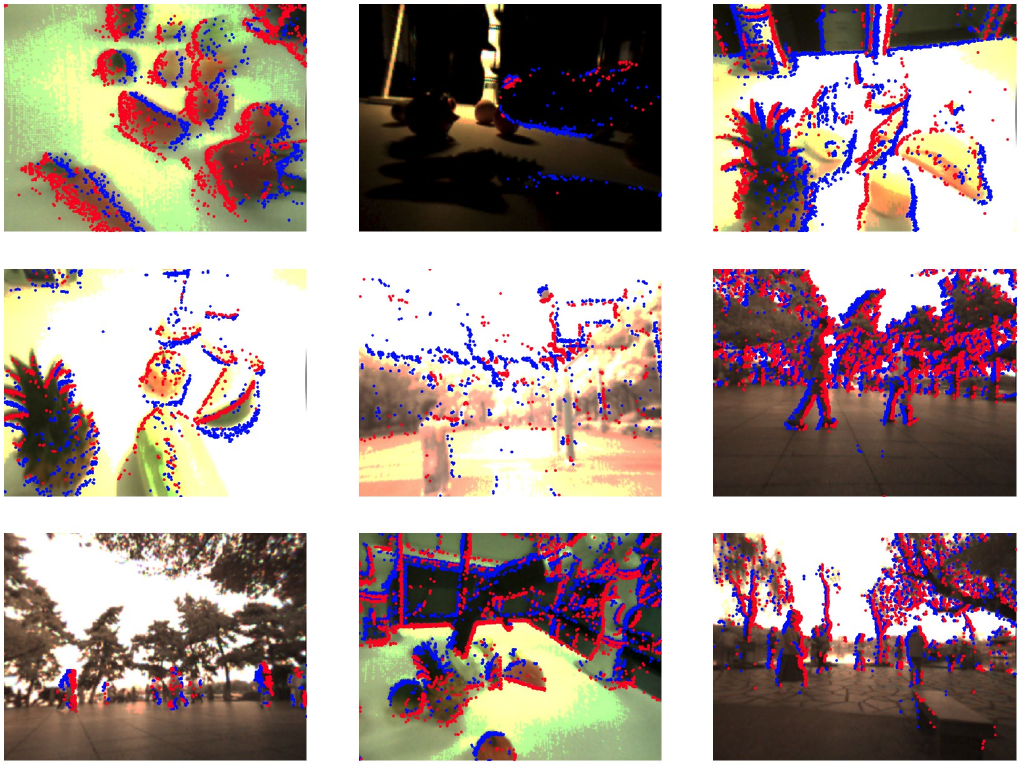} 
   \vspace{-0.1cm}
   \caption{A few examples of InivTAP. }
   \vspace{-0.5cm}
   \label{fig:InivData}
\end{figure}

\begin{figure*}[t]
   \centering
   \includegraphics[width=1\textwidth]{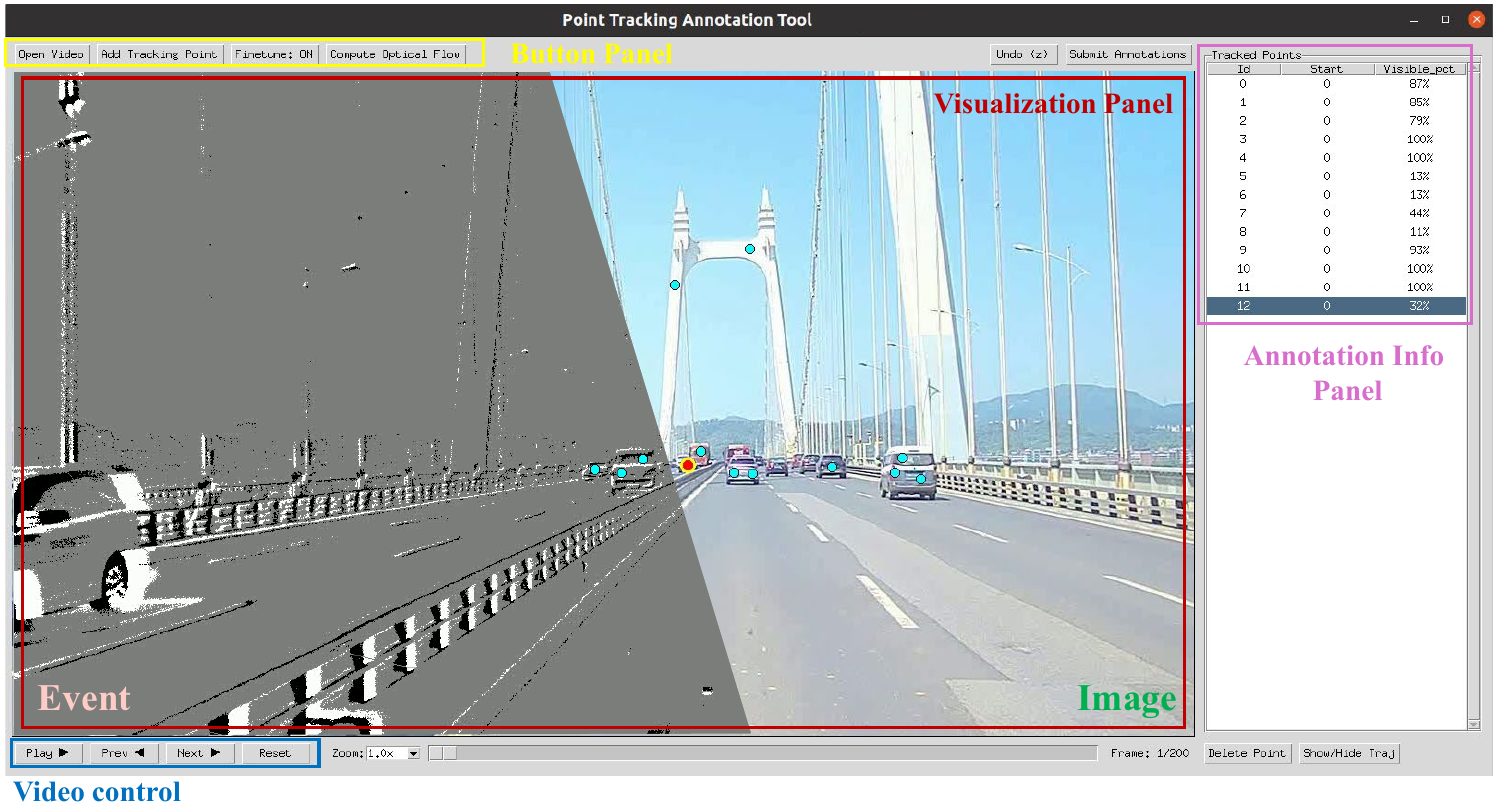} % 替换为你的图片文件名
   \caption{Annotation interface for the TAP task. We select either RGB frames or event-reconstructed frames for manual ground-truth labeling, depending on which modality provides clearer visual cues. The tool consists of three main components: a visualization panel for displaying the sequence, an annotation information panel for managing point information, and a button panel providing operation interactive controls. }
   \label{fig:annotation_tool}
\end{figure*}

\paragraph{InivTAP and DrivTAP}
In addition to the synthetic training set, we introduce two real-world frame–event TAP test sets, both manually annotated with point trajectories and occlusion labels. The experimental data collection equipment is shown in ~\figref{fig:device}, and the ground truth annotation tool is shown in ~\figref{fig:annotation_tool}. These datasets serve as reliable benchmarks for evaluating asynchronous frame–event tracking performance in real scenes, shown in ~\figref{fig:InivData}.

Each dataset follows a three-stage annotation pipeline to ensure trajectory accuracy: 
(1) Initial motion trajectories are generated using RAFT~\cite{gehrig2021raft} optical flow. 
(2) Human annotators refine each trajectory by referencing both the image frames and the event-reconstructed frames (obtained using E2Vid~\cite{rebecq2019events}); the annotation is performed on the modality that offers clearer visual cues. 
(3) High-frequency noise is filtered to obtain the final smooth ground-truth trajectories.

For point selection, we prioritize targets that (i) remain visible for long durations, (ii) exhibit noticeable motion, and (iii) distributed across different moving objects, with no more than three points per object (five if one object moves).

\begin{figure*}[t]
   \centering
   \includegraphics[width=1\textwidth]{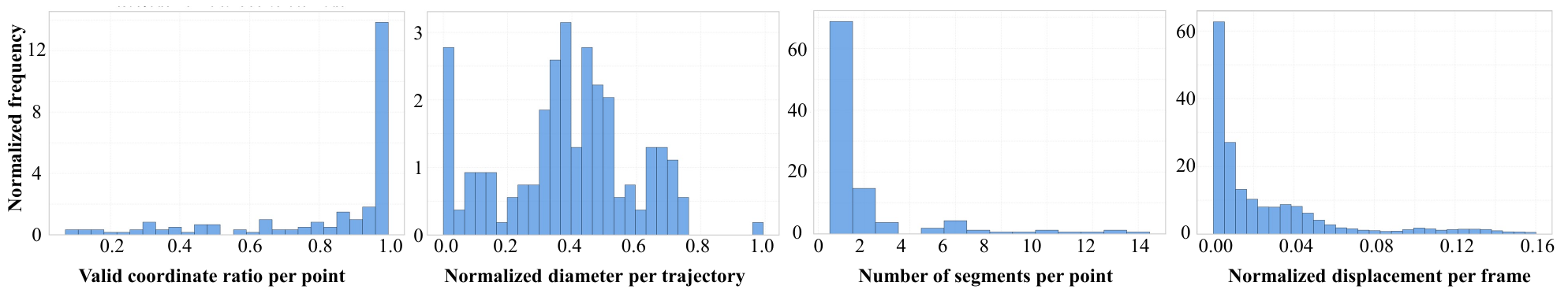} % 替换为你的图片文件名
   \caption{Statistics of ground-truth trajectories in InivTAP and DrivTAP. Valid coordinate ratio measures the percentage of time a queried point remains within the image boundaries. Diameter denotes the trajectory’s motion range normalized by image height. Number of segments indicates how many continuous track fragments are formed due to occlusions. Displacement represents the per-frame motion magnitude normalized by image height. }
   \label{fig:dataset_analyze}
\end{figure*}

Below, we summarize several key characteristics of our manually annotated test datasets. As shown in ~\figref{fig:dataset_analyze}, most point trajectories span nearly the entire video, with over 60\% of tracks present in more than 90\% of the sequence duration (though some may be occluded). In total, 80.04\% of points are never occluded, and most trajectories consist of a single continuous unoccluded segment.

Our datasets also exhibit substantial motion diversity. A notable portion of points undergo large displacements: 16.51\% move more than 5\% of the image height between consecutive frames, and 8.1\% exceed 10\%, indicating significant motion dynamics. Meanwhile, another portion of points remains nearly static, further highlighting the diversity of motion patterns covered by our dataset.

In addition, the test sequences include challenging conditions such as overexposure and low-light scenes, enabling a comprehensive evaluation of tracking performance across a broad range of real-world environments.

\begin{table*}[t]
\centering
\caption{Tracking performance comparison on the EC dataset. The first four rows list frame-only methods, followed by five event-only baselines, and the last four rows present approaches fusing frames and events. Methods with a gray background denote our models. All results are reported in percentages, and the best performance in each column is highlighted in \textbf{bold}.}
\label{tab:ec_full_result}
\renewcommand{\arraystretch}{1.3}
\resizebox{\textwidth}{!}{
\begin{tabular}{l cc cc cc cc cc cc}
\hline
\multirow{2}{*}{Method} & \multicolumn{2}{c}{Average} & \multicolumn{2}{c}{shapes\_trans} & \multicolumn{2}{c}{shapes\_rot} & \multicolumn{2}{c}{shapes\_6dof} & \multicolumn{2}{c}{boxes\_trans} & \multicolumn{2}{c}{boxes\_rot} \\
\cmidrule(lr){2-3} \cmidrule(lr){4-5} \cmidrule(lr){6-7} \cmidrule(lr){8-9} \cmidrule(lr){10-11} \cmidrule(lr){12-13}
& FA$\uparrow$ & EFA$\uparrow$ & FA$\uparrow$ & EFA$\uparrow$ & FA$\uparrow$ & EFA$\uparrow$ & FA$\uparrow$ & EAF$\uparrow$ & FA$\uparrow$ & EFA$\uparrow$ & FA$\uparrow$ & EFA$\uparrow$ \\
\hline
PIPs++~\cite{zheng2023pointodyssey} & 82.6 & 82.3 & 87.1 & 87.1 & 79.0 & 78.8 & 83.7 & 82.7 & 86.3 & 86.1 & 77.0 & 76.9 \\
Cotracker3~\cite{karaev2025cotracker3} & 92.5 & 91.9 & 94.4 & 94.3 & 87.3 & 86.8 & 93.3 & 91.6 & \textbf{93.6} & \textbf{93.2} & 94.3 & 93.9 \\
Chrono~\cite{kim2025exploring} & 82.9 & 82.3 & 85.1 & 84.5 & 81.5 & 81.1 & 75.6 & 74.1 & 83.3 & 83.0 & 88.8 & 88.6 \\
\rowcolor{gray!20} TAPFormer-F & 92.7 & 92.1 & 95.2 & 95.1 & 87.8 & 87.4 & \textbf{92.8} & \textbf{91.1} & 92.9 & 92.4 & 94.7 & 94.4 \\
\hline
EM-ICP~\cite{conf_icra_ZhuAD17} & 33.7 & 33.4 & 40.3 & 40.2 & 32.0 & 32.0 & 24.8 & 24.2 & 35.5 & 35.4 & 35.6 & 34.9 \\
HASTE~\cite{conf_bmvc_AlzugarayC20} & 44.2 & 42.7 & 58.9 & 56.4 & 61.3 & 58.2 & 13.3 & 4.3 & 38.2 & 36.8 & 49.2 & 44.7 \\
ETAP~\cite{hamann2025etap}  & 88.1 & 87.6 & 91.7 & 91.5 & 87.1 & 86.9 & 90.5 & 88.6 & 85.0 & 84.9 & 86.0 & 85.9 \\
MATE~\cite{han2025mate} & 88.5 & 87.5 & - & - & - & - & - & - & - & - & - & - \\
\rowcolor{gray!20} TAPFormer-E & 86.6 & 86.1 & 93.3 & 92.9 & 81.8 & 81.6 & 91.4 & 89.8 & 89.0 & 88.8 & 77.5 & 77.3 \\
\hline
EKLT~\cite{journals_ijcv_GehrigRGS20} & 81.1 & 77.5 & 83.9 & 74.0 & 83.3 & 80.6 & 81.7 & 69.6 & 68.2 & 64.4 & 88.3 & 86.5 \\
DeepEvT~\cite{conf_cvpr_MessikommerFG023} & 82.5 & 81.8 & 86.1 & 86.5 & 79.7 & 79.3 & 89.9 & 88.2 & 87.2 & 86.9 & 69.5 & 69.1 \\
FE-TAP ~\cite{liu2025tracking}& 84.4 & 83.8 & 93.1 & 92.9 & 81.5 & 81.3 & 87.9 & 86.0 & 73.1 & 72.8 & 86.2 & 86.1 \\
\rowcolor{gray!20} TAPFormer & \textbf{93.3} & \textbf{92.6} & \textbf{96.1} & \textbf{96.0} & \textbf{88.7} & \textbf{88.3} & \textbf{92.8} & \textbf{91.1} & 93.2 & 92.6 & \textbf{94.8} & \textbf{94.4} \\
\hline
\end{tabular}
}
\end{table*}

\begin{table*}[t]
\centering
\caption{Tracking performance comparison in EDS dataset. The values in parentheses indicate the performance of the Peanuts Light sequence under ground-truth evaluation before trajectory correction. For the Average column, the value in parentheses indicates the mean score after replacing only the Peanuts Light result. Rocket Earth$^*$ indicating that the ground truth is not fully accurate. All results are reported in percentages, and the best performance in each column is highlighted in \textbf{bold}.}
\label{tab:eds_full_result}
\renewcommand{\arraystretch}{1.3}
\resizebox{\textwidth}{!}{
\begin{tabular}{l cc cc cc cc cc}
\hline
\multirow{2}{*}{Method}  & \multicolumn{2}{c}{Average} & \multicolumn{2}{c}{Peanuts Light} & \multicolumn{2}{c}{Rocket Earth*} & \multicolumn{2}{c}{Ziggy Arena} & \multicolumn{2}{c}{Peanuts Running} \\
\cmidrule(lr){2-3} \cmidrule(lr){4-5} \cmidrule(lr){6-7} \cmidrule(lr){8-9} \cmidrule(lr){10-11}
& FA$\uparrow$ & EFA$\uparrow$ & FA$\uparrow$ & EFA$\uparrow$ & FA$\uparrow$ & EFA$\uparrow$ & FA$\uparrow$ & EFA$\uparrow$ & FA$\uparrow$ & EFA$\uparrow$  \\
\hline
PIPs++~\cite{zheng2023pointodyssey} & 75.1(72.1) & 63.0(60.3) & 62.6(50.9) & 58.8(47.9) & 76.4 & 34.2 & 85.1 & 85.0 & 76.1 & 74.1 \\
cotracker3~\cite{karaev2025cotracker3} & 80.2(75.8) & 68.8(64.7) & 72.7(54.9) & 68.4(51.9) & 69.2 & 30.7 & 92.6 & 92.5 & 86.4 & 83.7 \\
\rowcolor{gray!20} TAPFormer-F & 80.9(76.4) & 69.1(64.9) & 74.7(56.7) & 70.1(53.3) & 71.5 & 31.5 & 91.8 & 91.8 & 85.6 & 82.8 \\
\hline
EM-ICP~\cite{conf_icra_ZhuAD17} & 16.1 & 12.0 & 8.4 & 7.7 & 29.8 & 15.8 & 15.3 & 14.9 & 10.8 & 9.5 \\
HASTE~\cite{conf_bmvc_AlzugarayC20}  & 9.6 & 6.3 & 8.6 & 7.6 & 16.2 & 8.5 & 8.2 & 5.7 & 5.4 & 3.3 \\
ETAP~\cite{hamann2025etap}   & 74.5(70.1) & 63.9(60.1) & 70.2(53.7) & 66.1(50.7) & 67.6 & 33.3 & 83.8 & 83.7 & 76.2 & 72.6 \\
MATE~\cite{han2025mate}   & (71.3) & (62.6) & - & - & - & - & - & - & - & - \\
\rowcolor{gray!20} TAPFormer-E & 76.8(72.4) & 64.5(60.4) & 67.7(50.2) & 63.7(47.4) & \textbf{80.0} & \textbf{37.2} & 81.3 & 81.1 & 78.1 & 75.9 \\
\hline
EKLT~\cite{journals_ijcv_GehrigRGS20}  & 32.5 & 20.5 & 28.4 & 26.0 & 42.5 & 17.5 & 41.9 & 23.1 & 17.1 & 15.3 \\
DeepEvT~\cite{conf_cvpr_MessikommerFG023}  & 61.3(57.6) & 50.5(47.2) & 59.4(44.7) & 55.5(42.0) & 64.8 & 29.1 & 74.8 & 74.6 & 46.0 & 42.8 \\
FE-TAP~\cite{liu2025tracking}  & 72.2(67.6) & 63.2(58.9) & 73.1(54.9) & 68.9(51.7) & 53.8 & 24.6 & 84.9 & 84.4 & 76.9 & 74.9 \\
\rowcolor{gray!20} TAPFormer & \textbf{82.3(77.6)} & \textbf{70.4(66.1)} & \textbf{76.5(57.9)} & \textbf{71.5(54.4)} & 73.8 & 34.1 & \textbf{92.7} & \textbf{92.7} & \textbf{86.1} & \textbf{83.3} \\ 
\hline
\end{tabular}
}
\end{table*}

\begin{table*}[t]
\centering
\footnotesize
\setlength{\tabcolsep}{3.5pt}
\renewcommand{\arraystretch}{1.2}
\caption{Dataset comparison. Overview of publicly available synthetic event-based motion estimation datasets.}
\label{tab:datasets}
\begin{tabular}{l c c c c c c c c c c c c}
\toprule
\multirow{2}{*}{Dataset} &
\multirow{2}{*}{Source} &
\multirow{2}{*}{Events} &
\multirow{2}{0.7cm}{\centering Final \\ \centering image} &
\multirow{2}{0.7cm}{\centering Fast \\ \centering scene} &
\multirow{2}{*}{\#Samples} &
\multirow{2}{*}{Resolution} &
\multirow{2}{*}{fps[Hz]} &
\multirow{2}{1.3cm}{\centering sample \\ \centering duration[s]} &
\multicolumn{4}{c}{{Annotations}} \\
\cmidrule(lr){10-13}
 & & & & & & & & &
optical flow & TAP & depth & segmentations \\
\midrule

TAP-Vid Kubric\cite{doersch2022tap} &
3D PBR &
none & $\checkmark$ & $\times$ &
$\approx 10000$ &
$512 \times 512$ &
12 & 2 &
$\checkmark$ & $\checkmark$ & $\checkmark$ & $\checkmark$ \\

BlinkFlow\cite{li2023blinkflow} &
3D PBR &
synthetic & $\checkmark$ & $\times$ &
3587 &
$640 \times 480$ &
10 & 1 &
$\checkmark$ & $\times$ & $\checkmark$ & $\checkmark$ \\

MultiFlow\cite{gehrig2024dense} &
2D warp &
synthetic & $\times$ & $\times$ &
12100 &
$512 \times 384$ &
100 & 0.5 &
$\checkmark$ & $\times$ & $\times$ & $\times$ \\

EventKubric\cite{hamann2025etap} &
3D PBR &
synthetic & $\times$ & $\times$ &
10173 &
$512 \times 512$ &
48 & 2 &
$\checkmark$ & $\checkmark$ & $\checkmark$ & $\checkmark$ \\

% Repeat last row
\textbf{FE-FastKub(Ours)} &
3D PBR &
synthetic & $\checkmark$ & $\checkmark$ &
10953 &
$512 \times 512$ &
48 & 2 &
$\checkmark$ & $\checkmark$ & $\checkmark$ & $\checkmark$ \\

\bottomrule
\end{tabular}
\vspace{-0.2cm}
\end{table*}

\FloatBarrier
\clearpage
% \clearpage
{
    \small
    \bibliographystyle{ieeenat_fullname}
    \bibliography{main}
}


\end{document}